\documentclass{article}

\usepackage{iclr2027_conference,times}

\usepackage{amsmath,amsfonts,bm}

\def\eqref#1{equation~\ref{#1}}

\def\1{\bm{1}}

\DeclareMathAlphabet{\mathsfit}{\encodingdefault}{\sfdefault}{m}{sl}
\SetMathAlphabet{\mathsfit}{bold}{\encodingdefault}{\sfdefault}{bx}{n}

\usepackage{hyperref}
\usepackage{url}
\usepackage{graphicx}
\usepackage{float}
\usepackage{booktabs}
\usepackage{listings}
\usepackage{algorithm}
\usepackage{algpseudocode}
\usepackage{xcolor}
\usepackage{colortbl}
\definecolor{resightTint}{RGB}{233,245,241}
\definecolor{resightGray}{RGB}{246,246,246}
\definecolor{resightOrange}{RGB}{255,239,219}

\title{ReSight: Task-Driven Self-Improving Memory for Embodied Agents}

\author{%
\begin{minipage}[t]{\dimexpr\textwidth-2\tabcolsep\relax}
\centering\normalfont\normalsize
Lirong Che\textsuperscript{1,2}, Yang Chen\textsuperscript{2,3}, Yimeng Su\textsuperscript{1,2}, Daqi Liu\textsuperscript{2},\\
Zhenyu Huang\textsuperscript{2,3}, Jiong Wang\textsuperscript{2,4}, Yuzhe Yang\textsuperscript{2}, Xu Cao\textsuperscript{2},\\
Chuang Wang\textsuperscript{2}, Jian Su\textsuperscript{2}, Xueqian Wang\textsuperscript{1}\\[0.25em]
{\small
\textsuperscript{1}Tsinghua University\qquad
\textsuperscript{2}AGIBOT\\
\textsuperscript{3}Nanjing University\quad
\textsuperscript{4}Nanyang Technological University}
\end{minipage}%
}

\iclrfinalcopy
\hypersetup{
  pdftitle={ReSight: Task-Driven Self-Improving Memory for Embodied Agents},
  pdfauthor={Lirong Che, Yang Chen, Yimeng Su, Daqi Liu, Zhenyu Huang, Jiong Wang, Yuzhe Yang, Xu Cao, Chuang Wang, Jian Su, Xueqian Wang}
}

\begin{document}

\vspace*{-0.5em}
\maketitle
\pagestyle{plain}
\thispagestyle{plain}
\vspace{-1.5em}

\begingroup
\setlength{\intextsep}{4pt}
\setlength{\abovecaptionskip}{4pt}
\begin{figure}[H]
\centering
\includegraphics[width=\linewidth]{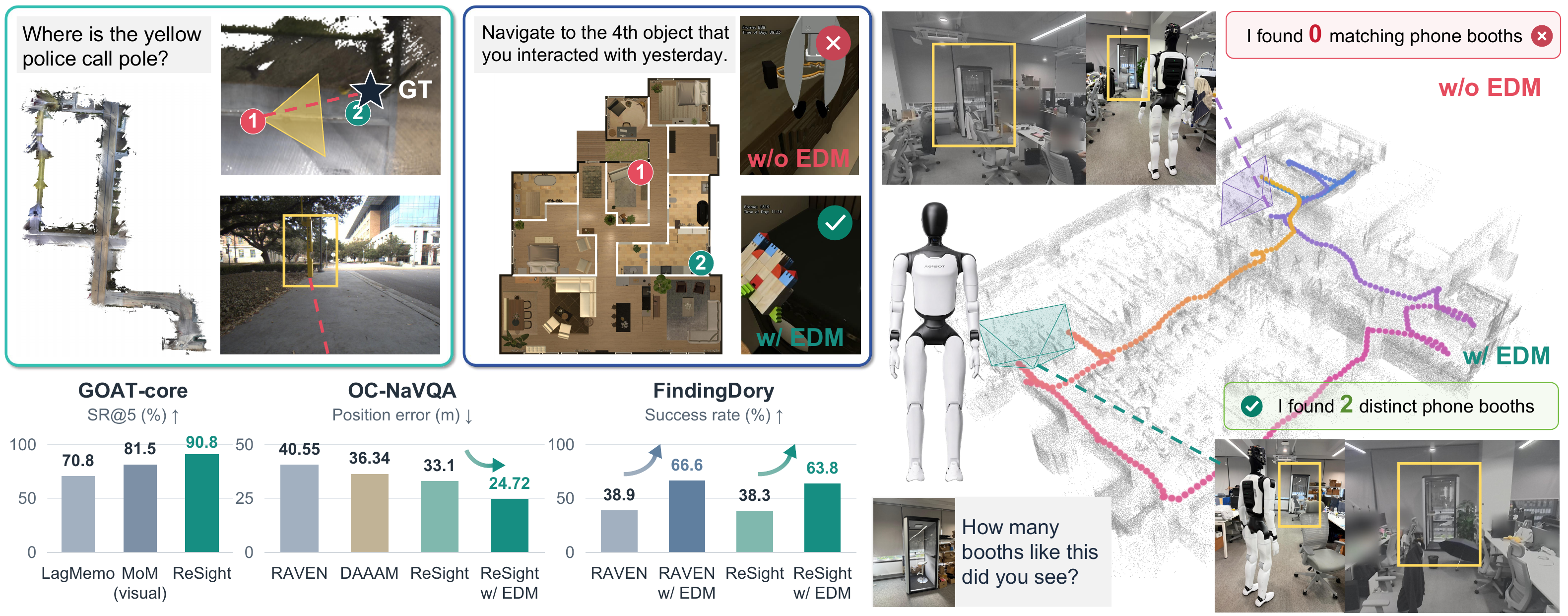}
\caption{ReSight retains revisitable visual evidence and expands its processing capabilities through EDM. Evaluation across scan-based indoor simulation (GOAT-core), interactive household simulation (FindingDory), real campus recordings (OC-NaVQA), and real-robot office patrols demonstrates effective memory retrieval and task-driven capability expansion.}
\label{fig:resight_overview}
\end{figure}
\endgroup

\vspace{-0.4em}
\begin{abstract}
Robots can use historical observations to answer spatiotemporal queries in familiar environments, but the presence of relevant evidence does not imply that existing tools support the operations a task requires. We introduce ReSight, a task-driven self-improving memory system for embodied agents that expands the capabilities for processing and accessing historical observations as task demands arise. ReSight retains keyframes and their spatiotemporal associations, separating observation evidence from extensible derived fields so that stored observations can be reprocessed for new needs. An execution agent and a coding agent share this memory through different interfaces: the former calls registered tools within a bounded online budget, while the latter uses a programmable workspace and a larger exploration budget to inspect evidence, write programs, and acquire models as needed. To realize capability expansion in ReSight, we develop Explore--Distill--Mount (EDM), which identifies capability gaps from execution and exploration records without task reference answers and converts exploration-derived solutions into reusable memory builders, query tools, and skills, extending memory representations and access capabilities as needed. Experiments on three public benchmarks demonstrate effective base memory access and capability expansion. With the execution model and online budget held fixed within each benchmark, EDM reduces mean position error on OC-NaVQA from 33.10\,m to 24.72\,m and improves high-level success on FindingDory from 38.3\% to 63.8\%. With the tools and skill frozen, success still improves by 28.3 percentage points on FindingDory episodes excluded from exploration and distillation. Experiments on real robot patrol memories further demonstrate reuse of these capabilities for subsequent queries and new memories, with fewer online tool calls on average.

\end{abstract}

\section{Introduction}
Long-horizon robot tasks such as fetching and delivery often begin with a decision about where to go to satisfy a request. In previously visited environments, past observations can inform this choice. Finding emergency equipment requires recalling where it was seen; finding an object that has moved also requires distinguishing observations from different times. Such place selection depends on spatiotemporal question answering over memory, whose outputs can guide subsequent navigation and manipulation. We study tasks whose relevant evidence is already present in retained observations: how can an agent find, interpret, and combine that evidence to answer the current query?

Embodied memory takes several forms, including scene graphs for structured queries and visual memories associated with time and position~\citep{resight_daaam,resight_raven}. Retained keyframes also allow object localization and representation construction to be deferred until query time~\citep{resight_mom,resight_foundit}. These approaches provide a basis for finding relevant observations, but the computation required by a task may exceed the capabilities of the available tools. Semantic retrieval may return images of a corridor, while distinguishing similar doors requires reading their signs. After retrieving several observations of an object, determining where it stayed longest requires organizing events across time. A task exposes a \emph{capability gap} when the execution agent cannot readily perform these operations with its existing tools and online budget.

We introduce ReSight, which combines revisitable spatiotemporal evidence with extensible memory capabilities. Unlike a fixed memory interface, ReSight can later acquire operations such as text reading, interaction-history reasoning, and object localization by reprocessing observations that are already stored. Its \emph{spatiotemporal evidence memory} retains selected keyframes and their temporal associations, together with available observations such as poses and depth. Scene and object descriptions, along with visual and text features, form initial semantic fields for retrieval. Subsequent needs can prompt new processing of stored images to derive text, event, or geometric fields while preserving their links to source observations. This \emph{late binding} allows interpretation and computation to evolve after observations have been stored, and lets memory builders and query tools be developed together for a new task requirement.

ReSight separates online response from capability development through two agents (Figure~\ref{fig:resight_architecture}). The execution agent treats the evidence memory as a database accessed through registered tools, using skills to guide retrieval, inspection, and answering within a bounded online budget. The coding agent treats the same memory as a programmable workspace: it can inspect files and images, write and execute code, and download and install models and dependencies. A separate, larger exploration budget allows it to develop solutions for requirements that are difficult for the current executor. The two agents work with the same evidence, carrying out bounded execution and open-ended exploration, respectively, providing a basis for turning exploration-derived solutions into reusable capabilities for the executor.

ReSight implements task-driven capability expansion through Explore--Distill--Mount (EDM). Prior work uses execution traces to improve programs, skills, and memory-management mechanisms~\citep{resight_aspire,resight_metaharness,resight_memevolve}. EDM combines execution-side capability gaps with coding-agent exploration traces to extract reusable tools, skills, and memory builders for retained sensor evidence. Builders reprocess stored observations to produce persistent fields and indices; corresponding query tools and skills are mounted on the execution agent so it can use the new capabilities.

Our contributions are as follows:
\begin{itemize}
    \item We introduce ReSight, which retains revisitable spatiotemporal observations and provides distinct access interfaces for execution and coding agents, allowing historical observations to be reprocessed for new task demands and memory representations and access capabilities to be extended as needed.
    \item We propose Explore--Distill--Mount (EDM), which identifies capability gaps from execution and exploration records without task reference answers and converts the coding agent's exploration-derived solutions into reusable tools, skills, and memory builders.
    \item We evaluate ReSight's base memory access, task-driven capability expansion, and frozen-capability reuse on three public benchmarks and real robot patrol memories. With the execution model and online budget fixed within each benchmark, EDM improves spatial localization on OC-NaVQA and question answering on FindingDory. With the tools and skill frozen, success still improves by 28.3 percentage points on FindingDory episodes excluded from exploration and distillation. Real-robot experiments further demonstrate reuse of these capabilities for subsequent queries and new memories.
\end{itemize}

\section{Related Work}
\subsection{Spatial Representations and Target Localization}
Open-vocabulary maps associate language-aligned features with spatial locations for text queries and target localization~\citep{resight_vlmaps,resight_openscene,resight_conceptfusion}.
Continuous semantic fields, multimodal feature memories, and 3D object instances offer alternative representations~\citep{resight_clipfields,resight_m3,resight_openmask3d}.
Scene graphs organize object relations and spatial hierarchies~\citep{resight_conceptgraphs,resight_hovsg}, while dynamic point clouds, spatiotemporal maps, and description histories capture environmental changes~\citep{resight_dynamem,resight_khronos,resight_daaam}.
Task-driven representations adjust semantic granularity according to predefined tasks or subsequent queries~\citep{resight_clio,resight_foundit}.

\subsection{Long-Horizon Visual Memory and Retrieval}
Observation descriptions, event records, and object tracks can form queryable long-term memories~\citep{resight_remembr,resight_embodiedrag,resight_videoagent}; multi-view snapshots, visual embeddings, and keyframe-augmented scene graphs retain visual context~\citep{resight_threedmem,resight_raven,resight_keysg}.
Semantic-spatial retrieval and memory fusion further select and integrate relevant information~\citep{resight_metamemory,resight_threedllmmem}.
Recorded tours can supply goal frames for navigation~\citep{resight_mobilityvla}, and retrieved RGB-D keyframes support on-demand target localization~\citep{resight_mom}; prediction error can guide which events are stored~\citep{resight_worth}.
ReSight retains sensor observations as reprocessable evidence and adds structured fields as subsequent needs arise, allowing memory processing and access capabilities to expand.

\subsection{Programmatic Problem Solving and Harness Evolution}
Programmatic agents compose perception, control, and computation through code~\citep{resight_cap,resight_viper}, amortize generation costs through tool reuse~\citep{resight_latm}, and accumulate skills from execution feedback~\citep{resight_voyager,resight_aspire}.
Automatic improvement spans prompts and experiential contexts~\citep{resight_gepa,resight_ace}, as well as agent programs and their organization~\citep{resight_adas,resight_sica,resight_dgm}; synthesized code can also constrain action selection or implement policies directly~\citep{resight_autoharness}.
Recent harness evolution draws improvement signals from execution traces and demonstrations~\citep{resight_metaharness,resight_ahe,resight_selfharness,resight_demoevolve}, updating tools, runtime mechanisms, and component organization~\citep{resight_harnessx,resight_harnesscompass}, as well as memory construction and retrieval programs~\citep{resight_memevolve}.
EDM synthesizes tools, skills, and memory builders from execution-side capability gaps and coding-agent exploration traces, extending retained sensor evidence with spatiotemporal fields and access capabilities reusable during online execution.

\section{Method}
\begin{figure}[ht]
    \centering
    \includegraphics[width=\linewidth]{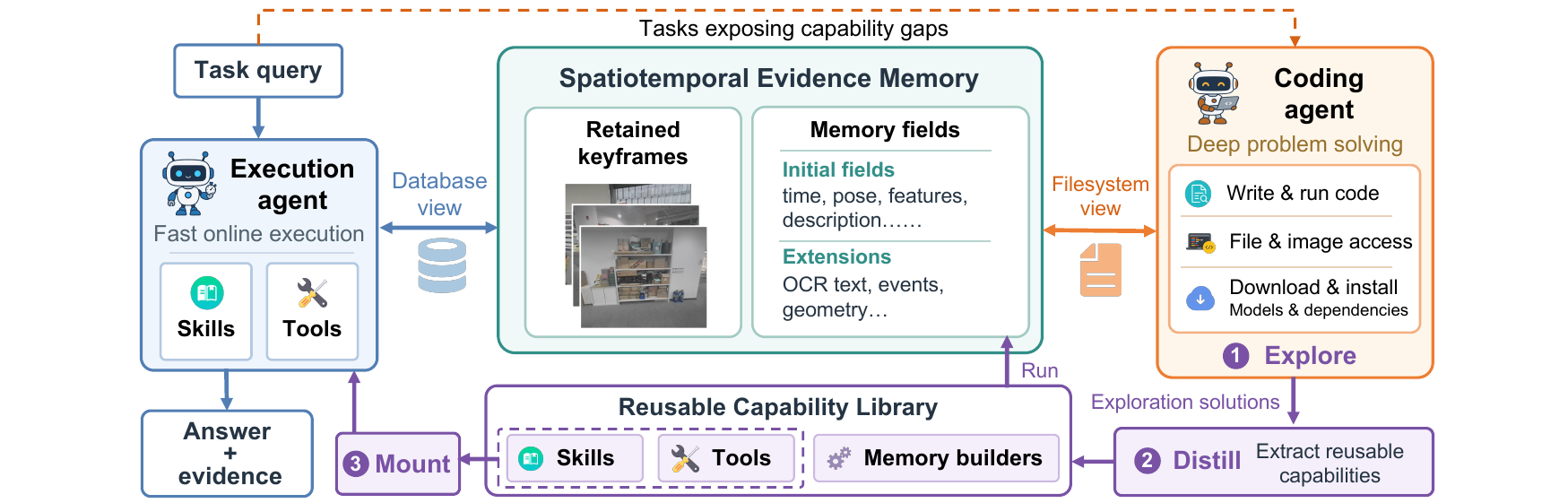}
    \caption{ReSight architecture. The two agents access the same evidence memory through different interfaces. EDM converts exploration-derived solutions into tools, skills, and memory builders for reuse by the execution agent.}
    \label{fig:resight_architecture}
\vspace{-4pt}
\end{figure}

ReSight provides a shared spatiotemporal evidence memory with extensible derived fields, accessed through asymmetric interfaces for execution and coding agents. EDM implements capability expansion within this architecture (Figure~\ref{fig:resight_architecture}).

\subsection{Problem Setting}
Given the robot's accumulated observation memory $\mathcal M$ and a task query $q$, the system returns an answer $y$ and supporting evidence $E_q$. Task-location queries additionally require a position or candidate place suitable for navigation. We consider tasks whose required information is present in retained observations. A task exposes a \emph{capability gap} when the current tools, skills, or memory-access mechanisms cannot solve it reliably within the online budget. For example, a room sign may be visible in memory, yet searching images individually can exhaust the call budget, motivating dedicated text extraction and retrieval.

\subsection{Spatiotemporal Evidence Memory}
ReSight anchors memory in retained RGB keyframes and their associated observations and acquisition context. Available data vary with the robot's sensors and the benchmark, including temporal associations, depth, poses, calibration, and synchronization information. We preserve these records at ingestion, especially information that is difficult to recover reliably from images, for subsequent processing. Let $\mathcal O$ denote the retained records, $\mathcal F^{(0)}$ the initial fields, and $\mathcal X^{(r)}$ the derived fields and indices accumulated after $r$ extensions. The memory is
\begin{equation}
    \mathcal M^{(r)}=(\mathcal O,\mathcal F^{(0)},\mathcal X^{(r)}).
    \label{eq:resight_memory}
\end{equation}

Initial fields record or reference observation times and available poses, and include descriptions and features produced during memory construction. A VLM generates scene and object descriptions for selected keyframes without conditioning on subsequent queries. These capture global scene semantics and object categories, attributes, and spatial context, respectively. We encode both types of description into text embeddings and the images into visual embeddings. Fields are linked to images by frame identifiers; events spanning multiple frames refer to the corresponding observation sets. Observation poses record camera locations, while object positions can be computed from depth or multiple views.

Text queries retrieve candidates independently from visual, scene-description, and object-description embeddings, then combine them by reciprocal rank fusion (RRF). Image queries use the visual channel. Appendix~\ref{app:retrieval-fusion} details fusion, filtering, and candidate grouping.

Subsequent tasks can introduce fields for OCR, events, or geometry. \emph{Memory builders} generate these fields and their query indices from retained observations, and new tools expose the operations required by a task. We use \emph{late binding} to describe determining how observations are interpreted and processed after storage: new programs can reprocess retained records to extract information absent from initial fields. \emph{Initial} and \emph{Extensions} distinguish when fields are introduced, independently of their source. Depth, for example, may be supplied by a sensor or dataset, or estimated from retained images.

\subsection{Asymmetric Access and Dual-Agent Problem Solving}
The execution and coding agents share the same memory but have different interfaces and compute budgets. The execution agent receives a \emph{database view} through registered tools and uses skills to guide retrieval, verification, and answering. Retrieval initially returns compact information, including frame identifiers, timestamps, locations, and descriptions. The agent requests images when it needs to verify details. This progressive disclosure reduces the context consumed by candidate screening under a bounded online call budget.

The coding agent receives a programmable \emph{workspace view}. It can search and read files and images, write, execute, and debug programs, and download models or install dependencies as needed. Exploration has a compute budget independent of the execution agent's per-query call limit, accommodating model preparation, memory processing, and debugging. The coding agent can reuse existing capabilities or construct new procedures, returning an answer while recording its solution, evidence, and exploration trace. EDM then transfers reusable parts to the execution agent.

\subsection{Explore--Distill--Mount}
\begin{figure}[t]
\centering
\includegraphics[width=\linewidth]{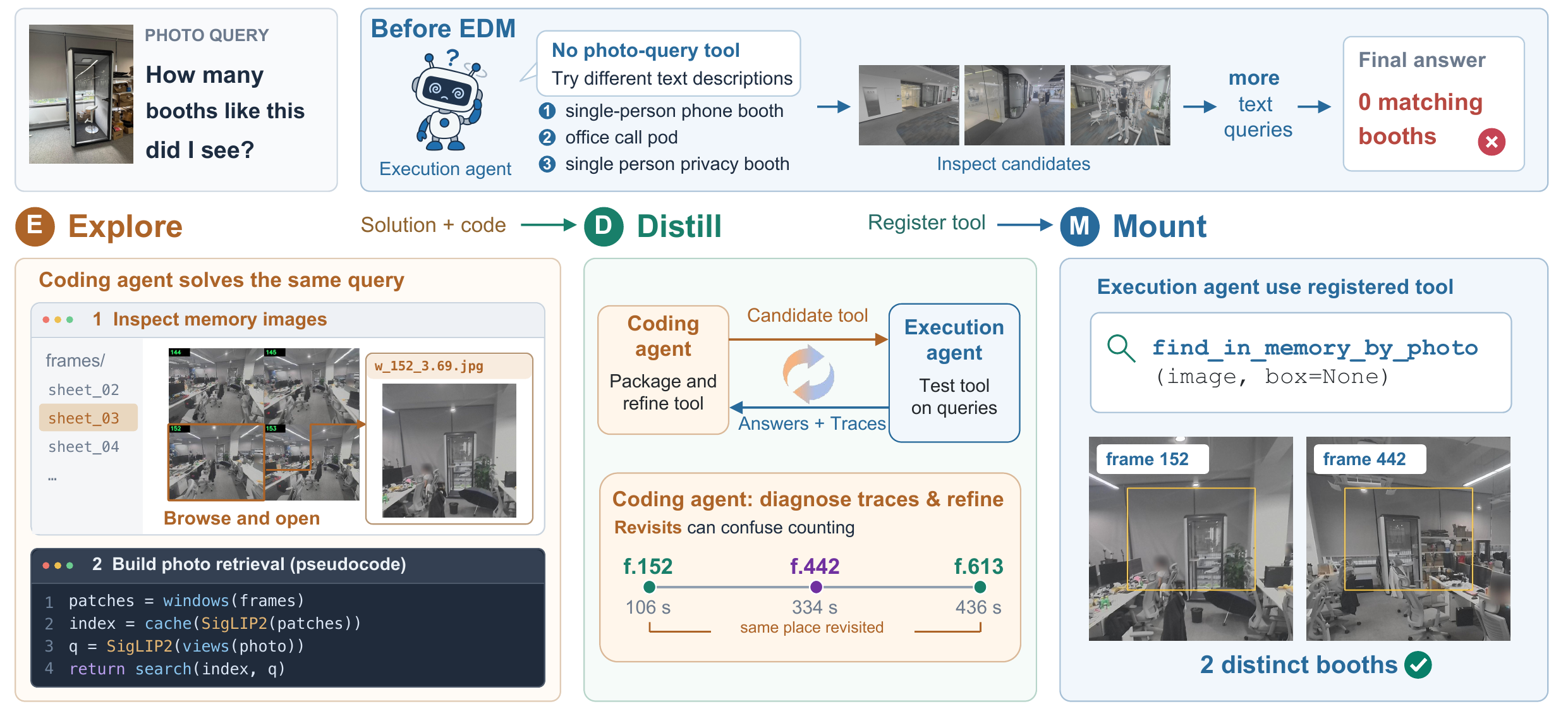}
\caption{EDM on robot-memory booth counting. After text-based retrieval fails, the coding agent develops photo retrieval and verifies two booths. Executor replay during distillation reveals that revisits confuse counting, prompting pose-and-time merging; the mounted tool returns both booths and supporting frames in one call.}
\label{fig:edm-booth}
\vspace{-4pt}
\end{figure}

EDM converts execution traces and exploration traces into capability-library updates, implementing demand-driven harness evolution. Let $H$ denote the execution configuration, including callable tools, skills that guide their use, and memory builders that produce persistent fields and indices. With retained records $\mathcal O$ and the execution model fixed, let $s_B(q;H)$ denote the expected task score under configuration $H$ and online budget $B$, with higher scores indicating better performance. EDM's design objective is to improve performance on capability-gap tasks:
\begin{equation}
    \max_{\Delta H}\;\mathbb E_{q\sim\mathcal D_{\mathrm{gap}}}
    \!\left[s_B(q;H')-s_B(q;H)\right],
    \qquad H'=\operatorname{Mount}(H,\Delta H).
    \label{eq:resight_objective}
\end{equation}
Here $\Delta H$ is the capability extension, $H'$ the mounted configuration, and $\mathcal D_{\mathrm{gap}}$ the distribution of tasks requiring capability expansion. EDM extracts reusable operations from exploration traces and integrates them with the host. Figure~\ref{fig:edm-booth} follows one task, counting phone booths from a reference photo, through the three steps.

\paragraph{Explore: develop a solution.}
The coding agent receives an evidence memory with all queries posed on it and solves them jointly in one session. It inspects images, invokes models, and writes programs, returning an answer to every query with the programs, notes, and evidence behind it. New fields may first appear at this stage. For example, the coding agent can run OCR to locate a room sign and verify the extracted text against the image. Exploration and distillation use observations and exploration traces without task reference answers.

\paragraph{Distill: extract reusable capabilities.}
Exploration may interleave scripts, shell commands, direct file inspection, and decisions based on intermediate results. Distillation turns reusable parts of this process into operations callable through the execution agent's registered interface. Failed or inefficient runs in the executor's traces locate the difficulties to address. The coding agent reads the executor's tool inventory and these traces, reuses existing capabilities, and packages missing or partially supported operations as tools with explicit inputs and outputs. Reusable preprocessing is packaged as memory builders.

\paragraph{Mount: integrate with execution.}
The distilled tools are registered with the execution agent. When a skill is mounted, the executor receives it in its system prompt: a short text on when and how to use the tools. Builder-generated fields and indices are exposed through query tools. For a new builder $b$ with output $\mathcal X_b$, the memory update is
\begin{equation}
    \mathcal X_b=b(\mathcal M^{(r)}),\qquad
    \mathcal M^{(r+1)}=(\mathcal O,\mathcal F^{(0)},\mathcal X^{(r)}\cup\mathcal X_b).
    \label{eq:resight_extension}
\end{equation}
Model initialization and whole-memory processing can precede mounting; query-dependent computation runs when a tool is called. Builders can be rerun when memory data change. Subsequent tasks use the same execution model and online budget with the updated capabilities.

\section{Experiments}
\providecommand{\resightchange}[1]{{\scriptsize\color{black!65}$(#1)$}}

We evaluate base memory access on GOAT-core, capability expansion on OC-NaVQA and FindingDory, and reuse on new queries and memories in real-robot patrol experiments.

\subsection{Experimental Setup}
ReSight denotes the complete system; w/o EDM and w/ EDM compare the same executor before and after capability mounting. We also evaluate RAVEN as a second host. Within each OC-NaVQA and FindingDory comparison, GPT-5.6-sol (Sol) and the online decision-round cap remain fixed. Opus-5 coding-agent results provide an exploration reference under a separate compute budget. Retrieval settings appear in Appendix~\ref{app:findingdory}.

\subsection{Base Memory Design on GOAT-core}
We evaluate ReSight's initial memory fields and multi-channel retrieval against MoM's SigLIP2-only visual-retrieval configuration.

Following the MoM protocol, we evaluate image, object, and language target queries across four GOAT-core scenes, for a total of 480 queries~\citep{resight_mom}. All reproduced configurations share the localization backend and return up to five candidates per query. We report SR@5 with a 1.5\,m success threshold on the navigation plane. ReSight's retrieval configuration returns candidates directly, while its agent configuration can perform further retrieval and inspect images.

\begin{table}[t]
\centering
\caption{GOAT-core SR@5 (\%); no EDM. The first six rows are from MoM Table~I~\citep{resight_mom}; HOV-SG averages two supported query types. Retrieval-only variants use no query-time VLM reasoning. Qwen/Sol denote annotation models (Appendix~\ref{app:goat-models}); bold marks group bests.}
\label{tab:goatcore}

\setlength{\tabcolsep}{5pt}
\begin{tabular}{@{}lrrrr@{}}
\toprule
\textbf{Method} & \textbf{Image} $\uparrow$ & \textbf{Object} $\uparrow$ & \textbf{Language} $\uparrow$ & \textbf{Average} $\uparrow$ \\
\midrule
DynaMem~\citep{resight_dynamem} & 16.1 & 7.5 & 5.1 & 9.6 \\
VLMaps~\citep{resight_vlmaps} & 43.3 & 69.7 & 61.0 & 58.8 \\
LagMemo~\citep{resight_lagmemo} & 56.4 & 88.4 & 66.8 & 70.8 \\
OpenFunGraph~\citep{resight_openfungraph} & 53.4 & 67.1 & 69.0 & 63.1 \\
HOV-SG~\citep{resight_hovsg} & -- & 70.8 & 82.3 & 76.5 \\
MoM~\citep{resight_mom} & 80.1 & 79.5 & 82.3 & 80.6 \\
ReSight (agent) & \textbf{92.5} & \textbf{90.7} & \textbf{88.6} & \textbf{90.6} \\
\midrule
\multicolumn{5}{l}{\textit{Retrieval-only variants}} \\
\addlinespace[2pt]
MoM (visual-only) & 88.2 & 72.7 & 83.5 & 81.5 \\
ReSight (retrieval, Qwen) & \textbf{90.1} & \textbf{91.3} & 88.0 & 89.8 \\
ReSight (retrieval, Sol) & \textbf{90.1} & 90.1 & \textbf{92.4} & \textbf{90.8} \\
\bottomrule
\end{tabular}
\vspace{-8pt}
\end{table}

Direct retrieval achieves 90.8\% SR@5 with Sol annotation and 89.8\% with Qwen, above MoM's visual-only 81.5\%; the agent achieves a similar 90.6\% (Table~\ref{tab:goatcore}). These results support the effectiveness of the initial memory and retrieval design.

\subsection{Capability Expansion with EDM}
We assess EDM on OC-NaVQA and FindingDory through before-and-after comparisons, additionally distilling and mounting capabilities for RAVEN as a second host on FindingDory.

\begin{figure}[t]
\centering
\includegraphics[width=\linewidth]{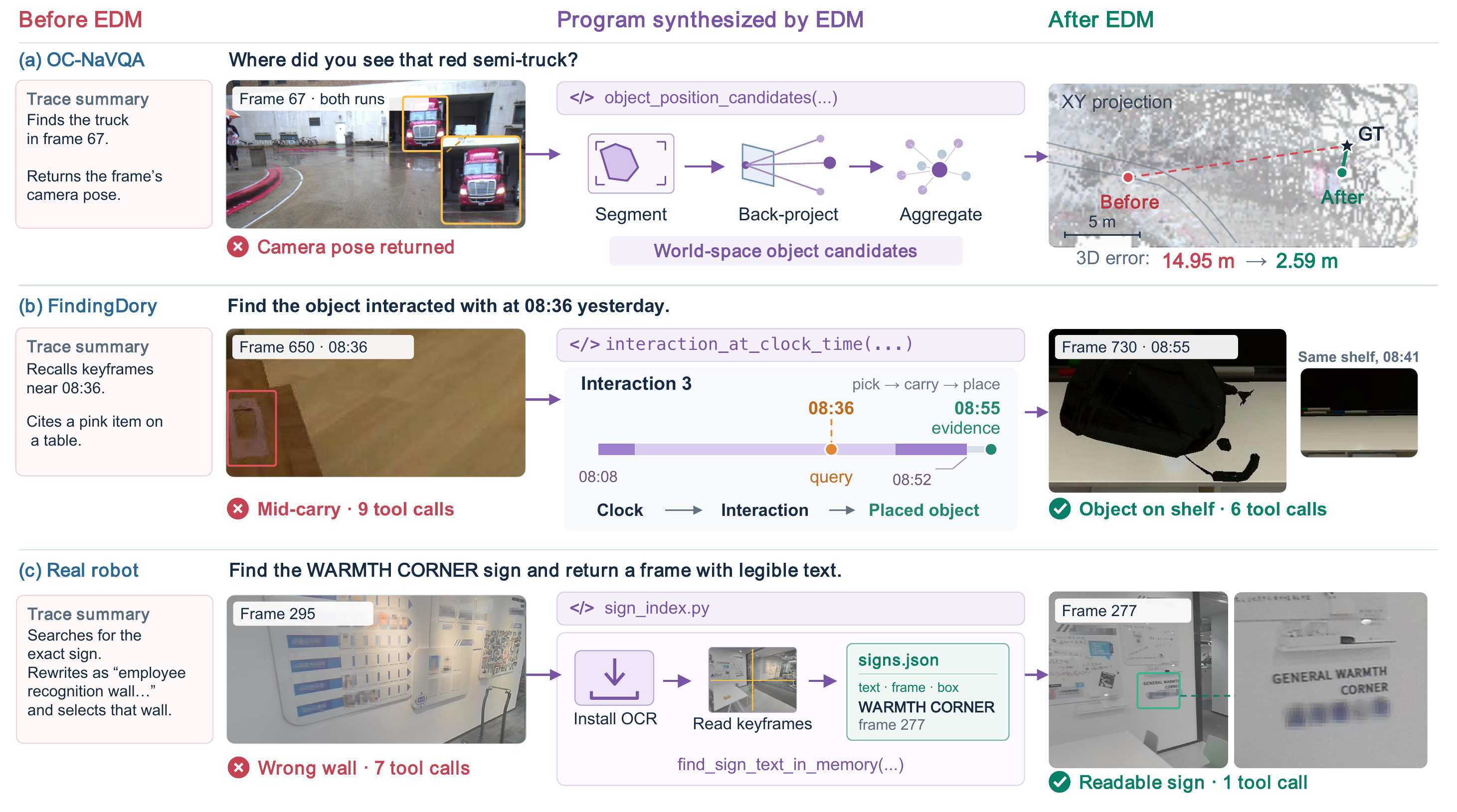}
\caption{Examples of EDM capability expansion: (a) object localization, (b) interaction-object tracking, and (c) OCR sign retrieval. Columns show execution before mounting, the generated programs, and execution after mounting. Call counts are per illustrated run; full trajectories appear in Appendix~\ref{app:case-records}.}
\label{fig:edm-cases}
\vspace{-4pt}
\end{figure}

\paragraph{OC-NaVQA.}
This benchmark contains 210 questions from seven sequences, covering binary, text, position, and temporal queries~\citep{resight_daaam}. Position questions require object coordinates, whereas ReSight's base executor returns an observation's camera pose. EDM adds object-position estimation among other capabilities; evaluation details appear in Appendix~\ref{app:ocnavqa}.

\begin{table}[t]
\centering
\caption{OC-NaVQA results on 210 questions. The first four rows are from DAAAM~\citep{resight_daaam}: Mini is GPT-5-mini; 2B/8B are NVILA-Lite models. RAVEN-Sol and ReSight-Sol$^\dagger$ return camera poses. Bold compares Sol configurations. Orange denotes the Opus-5 exploration reference.}
\label{tab:ocnavqa}

\setlength{\tabcolsep}{2.5pt}
\begin{tabular}{@{}lrrrrr@{}}
\toprule
& \multicolumn{2}{c}{Accuracy (\%)} & \multicolumn{2}{c}{Position} & Temporal \\
\cmidrule(lr){2-3}\cmidrule(lr){4-5}
\textbf{Method} & \textbf{Binary} $\uparrow$ & \textbf{Text} $\uparrow$ & \textbf{Mean (m)} $\downarrow$ & $\leq$\textbf{10\,m (\%)} $\uparrow$ & \textbf{Error (min)} $\downarrow$ \\
\midrule
ReMEmbR-2B~\citep{resight_remembr} & 43.2 & -- & 53.47 & -- & 2.287 \\
ReMEmbR-8B & 46.3 & -- & 55.89 & -- & 4.106 \\
ConceptGraphs~\citep{resight_conceptgraphs} & 29.9 & -- & 111.29 & -- & -- \\
DAAAM-Mini~\citep{resight_daaam} & 71.1 & -- & 41.75 & -- & 1.792 \\
\midrule
DAAAM-Sol & 80.6 & 57.6 & 36.34 & \textbf{52.1} & 2.225 \\
RAVEN-Sol~\citep{resight_raven} & 80.6 & 45.5 & 40.55 & 22.5 & 1.128 \\
\addlinespace[2pt]
ReSight-Sol (w/o EDM)$^\dagger$ & \textbf{86.6} & \textbf{63.6} & 33.10 & 19.7 & 0.587 \\
\rowcolor{resightTint}
ReSight-Sol (w/ EDM) & 82.1 & \textbf{63.6} & \textbf{24.72} & 36.6 & \textbf{0.475} \\
\midrule
\rowcolor{resightOrange}
CodingAgent-Opus-5 & 95.5 & 72.7 & 26.12 & 39.4 & 0.482 \\
\bottomrule
\end{tabular}
\vspace{-8pt}
\end{table}

EDM reduces mean position error from 33.10\,m to 24.72\,m and increases the fraction within 10\,m from 19.7\% to 36.6\% (Table~\ref{tab:ocnavqa}). The object-position tool combines image segmentation, depth backprojection, and multi-view aggregation to return target coordinates with supporting frames (Figure~\ref{fig:edm-cases}(a)).

\paragraph{FindingDory.}
We follow RAVEN's long-horizon subset, reporting official high-level success on 596 queries from episodes 91--100 across spatial, temporal, and multi-goal tasks~\citep{resight_raven}. ReSight and RAVEN each use a 15-decision-round budget. Capabilities are distilled separately for each host from exploration traces on these episodes. Appendix~\ref{app:findingdory} details retrieval and evaluation settings.

\begin{table}[t]
\centering
\caption{FindingDory high-level success (\%) on episodes 91--100 (596 queries). The first five rows are from RAVEN~\citep{resight_raven}; Gemini is Gemini-2.5-Flash. Bold compares Sol configurations; orange denotes the Opus-5 exploration reference. Tokens/q reports mean input-plus-output tokens in thousands. Annotation-model results appear in Appendix~\ref{app:annotator}.}
\label{tab:findingdory}

\setlength{\tabcolsep}{1.5pt}
\begin{tabular}{@{}lrrrrr@{}}
\toprule
\textbf{Method} & \textbf{Spatial} $\uparrow$ & \textbf{Temporal} $\uparrow$ & \textbf{Multi-goal} $\uparrow$ & \textbf{Overall} $\uparrow$ & \shortstack[r]{\textbf{Tokens/q}\\\textbf{(k)} $\downarrow$} \\
\midrule
Random & 13.4 & 6.9 & 0.5 & 9.2 & -- \\
Embedder Only & 31.0 & 8.1 & 3.4 & 18.8 & -- \\
VLM Only-Gemini & 36.0 & 28.1 & 2.2 & 28.2 & -- \\
ReMEmbR-Gemini~\citep{resight_remembr} & 36.7 & 17.1 & 4.5 & 25.0 & -- \\
RAVEN-Gemini~\citep{resight_raven} & 45.1 & 24.8 & 6.7 & 32.2 & -- \\
\midrule
RAVEN-Sol & 51.9 & 34.8 & 5.6 & 38.9 & 378 \\
\rowcolor{resightTint}
RAVEN-Sol (w/ EDM) & \textbf{55.9} & \textbf{89.5} & 48.3 & \textbf{66.6} & 150 \\
\addlinespace[2pt]
ReSight-Sol (w/o EDM) & 45.5 & 41.0 & 7.9 & 38.3 & 130 \\
\rowcolor{resightTint}
ReSight-Sol (w/ EDM) & 54.2 & 82.9 & \textbf{50.6} & 63.8 & \textbf{43} \\
\midrule
\rowcolor{resightOrange}
CodingAgent-Opus-5 & 58.2 & 76.2 & 43.8 & 62.4 & -- \\
\bottomrule
\end{tabular}
\vspace{-8pt}
\end{table}

EDM raises overall success from 38.3\% to 63.8\% on ReSight's executor and from 38.9\% to 66.6\% on RAVEN's, enabling both to reuse capabilities developed during exploration (Table~\ref{tab:findingdory}). After mounting, ReSight answers with 43k tokens per query against RAVEN's 150k: its retrieval returns text, and its added tools return compact text or composed evidence panels, while RAVEN places every retrieved frame in context as an image.

ReSight's executor gains most on multi-goal and temporal tasks, improving from 7.9\% to 50.6\% and from 41.0\% to 82.9\%. The distilled tools support interaction enumeration, revisit ordering, duration ranking, and clock-time alignment. In Figure~\ref{fig:edm-cases}(b), they link the query time to an interaction and return post-placement evidence, avoiding a frame captured while the object was still being carried.

\paragraph{Reuse on held-out episodes.}
We freeze the tools and skill distilled on episodes 91--100 and evaluate them on 597 queries from episodes 81--90, which were excluded from exploration and distillation, with no further capability updates. EDM raises overall success from 37.5\% to 65.8\% (+28.3 percentage points), while reducing tokens per query by 69.4\% (Table~\ref{tab:fd-heldout}). The gain is comparable to the 25.5-point gain on episodes 91--100, supporting reuse across episodes. Appendix~\ref{app:findingdory} gives the protocol. Distilling from the same execution traces without exploration programs, notes, or answers reaches 46.4\%: its event tool detects pauses of the robot base, which follow placements but miss pickups, and miscounts interactions in 3 of 10 episodes (Appendix~\ref{app:fd-wo-explore}). These results support the added value of exploration-derived solutions for building reusable temporal capabilities that generalize to held-out episodes.
\begin{table}[ht]
\centering
\caption{Frozen-capability reuse on FindingDory episodes 81--90 (597 queries). EDM w/o Explore distills from the execution traces (Appendix~\ref{app:fd-wo-explore}). High-level success (\%); Tokens/q reports mean input-plus-output tokens in thousands.}
\label{tab:fd-heldout}

\setlength{\tabcolsep}{4.8pt}
\begin{tabular}{@{}lrrrrr@{}}
\toprule
\textbf{Method} & \textbf{Spatial} $\uparrow$ & \textbf{Temporal} $\uparrow$ & \textbf{Multi-goal} $\uparrow$ & \textbf{Overall} $\uparrow$ & \shortstack[r]{\textbf{Tokens/q}\\\textbf{(k)} $\downarrow$} \\
\midrule
ReSight-Sol (w/o EDM) & 48.3 & 36.2 & 4.5 & 37.5 & 128.8 \\
ReSight-Sol (EDM w/o Explore) & 51.0 & 48.6 & 25.8 & 46.4 & \textbf{26.9} \\
\rowcolor{resightTint}
ReSight-Sol (w/ EDM) & \textbf{57.4} & \textbf{83.3} & \textbf{52.8} & \textbf{65.8} & 39.4 \\
\bottomrule
\end{tabular}
\vspace{-8pt}
\end{table}

\subsection{Real-Robot Experiments}
We evaluate ReSight on office-building patrol memories collected by an A3 robot, testing reuse of frozen programs on new queries and another floor. Exploration uses only floor-8 memory and three tasks: locating the Warmth Corner sign, finding the point farthest from the patrol start, and counting booths from a reference image. Both execution configurations use Sol with at most ten tool-interaction rounds.

We freeze the tools and builders, then evaluate the three exploration tasks and three new queries: room 805 on floor 8, and patrol length and photo localization on floor 1. The latter memory was collected on another day and excluded from exploration; frozen builders reconstruct its fields. Each task runs independently five times (Appendix~\ref{app:robot-evaluation}).

\begin{table}[t]
\centering
\caption{Real-robot memory queries, grouped into exploration and new tasks. Success is out of five runs; tool calls and input-plus-output tokens are per-run means. The last row aggregates 30 runs; bold marks pairwise bests. Queries and inputs: Appendix~\ref{app:full-queries}.}
\label{tab:real-robot}

\setlength{\tabcolsep}{4pt}
\begin{tabular}{@{}lrr rr rr@{}}
\toprule
& \multicolumn{2}{c}{\textbf{Success} $\uparrow$} & \multicolumn{2}{c}{\textbf{Tool calls} $\downarrow$} & \multicolumn{2}{c}{\textbf{Tokens/run (k)} $\downarrow$} \\
\cmidrule(lr){2-3}\cmidrule(lr){4-5}\cmidrule(lr){6-7}
\textbf{Task} & w/o EDM & w/ EDM & w/o EDM & w/ EDM & w/o EDM & w/ EDM \\
\midrule
Warmth Corner (8F) & 2/5 & \cellcolor{resightTint}\textbf{5/5} & 8.4 & \cellcolor{resightTint}\textbf{1.0} & 60.0 & \cellcolor{resightTint}\textbf{10.4} \\
Farthest point (8F) & 1/5 & \cellcolor{resightTint}\textbf{5/5} & 7.2 & \cellcolor{resightTint}\textbf{1.0} & 38.8 & \cellcolor{resightTint}\textbf{10.0} \\
Booth count (8F) & 0/5 & \cellcolor{resightTint}\textbf{5/5} & 7.6 & \cellcolor{resightTint}\textbf{1.0} & 50.2 & \cellcolor{resightTint}\textbf{12.9} \\
\midrule
Room 805 (8F) & \textbf{5/5} & \cellcolor{resightTint}\textbf{5/5} & 2.0 & \cellcolor{resightTint}\textbf{1.0} & 12.5 & \cellcolor{resightTint}\textbf{11.2} \\
Patrol length (1F) & 0/5 & \cellcolor{resightTint}\textbf{5/5} & 3.4 & \cellcolor{resightTint}\textbf{1.8} & \textbf{14.5} & \cellcolor{resightTint}15.1 \\
Photo localization (1F) & 4/5 & \cellcolor{resightTint}\textbf{5/5} & 2.0 & \cellcolor{resightTint}\textbf{1.4} & \textbf{11.7} & \cellcolor{resightTint}14.3 \\
\midrule
\textbf{Total / mean} & 12/30 & \cellcolor{resightTint}\textbf{30/30} & 5.10 & \cellcolor{resightTint}\textbf{1.20} & 31.29 & \cellcolor{resightTint}\textbf{12.31} \\
\bottomrule
\end{tabular}
\vspace{-8pt}
\end{table}

Mounting improves success from 12/30 to 30/30 and reduces mean online tool calls from 5.10 to 1.20 (Table~\ref{tab:real-robot}). The Warmth Corner text index also answers room 805; on the new floor, patrol-length success improves from 0/5 to 5/5 and photo localization from 4/5 to 5/5.

To isolate memory-related performance, we evaluate memory queries separately from downstream execution; Appendix~\ref{app:robot-execution} demonstrates how the retrieved locations support subsequent robot navigation, on-site inspection, and waiting at a user-specified destination.

\section{Limitations}
Our evaluation uses predefined query sets and task categories, enabling controlled comparisons of memory access and capability expansion. Real-world deployment can introduce new, long-tail queries as users' needs change. While our experiments demonstrate capability reuse across held-out episodes and new robot memories, they only partially capture this evolving task distribution. Evaluating repeated capability expansion over longer deployments and a broader range of emerging queries is an important direction for future work.

\section{Conclusion}
We introduced ReSight, which retains revisitable visual evidence and expands memory processing through EDM. Its dual-agent architecture separates bounded execution from programmatic exploration, distilling reusable tools, skills, and memory builders. Experiments demonstrate effective memory access, improved spatiotemporal question answering, and capability reuse across new queries and robot memories with fewer online tool calls.

\clearpage
\bibliography{resight_references}

@inproceedings{resight_vlmaps,
  title = {{Visual Language Maps for Robot Navigation}},
  author = {Huang, Chenguang and Mees, Oier and Zeng, Andy and Burgard, Wolfram},
  booktitle = {Proceedings of the IEEE International Conference on Robotics and Automation (ICRA)},
  year = {2023},
  pages = {10608--10615},
  url = {https://doi.org/10.1109/ICRA48891.2023.10160969}
}

@misc{resight_conceptgraphs,
  title = {{ConceptGraphs: Open-Vocabulary 3D Scene Graphs for Perception and Planning}},
  author = {Gu, Qiao and Kuwajerwala, Alihusein and Morin, Sacha and Jatavallabhula, Krishna Murthy and Sen, Bipasha and Agarwal, Aditya and Rivera, Corban and Paul, William and Ellis, Kirsty and Chellappa, Rama and Gan, Chuang and de Melo, Celso Miguel and Tenenbaum, Joshua B. and Torralba, Antonio and Shkurti, Florian and Paull, Liam},
  year = {2023},
  eprint = {2309.16650},
  archivePrefix = {arXiv},
  url = {https://arxiv.org/abs/2309.16650}
}

@inproceedings{resight_daaam,
  title = {{Describe Anything Anywhere At Any Moment}},
  author = {Gorlo, Nicolas and Schmid, Lukas and Carlone, Luca},
  booktitle = {Proceedings of the IEEE/CVF Conference on Computer Vision and Pattern Recognition (CVPR)},
  year = {2026},
  pages = {35002--35013},
  url = {https://openaccess.thecvf.com/content/CVPR2026/html/Gorlo_Describe_Anything_Anywhere_At_Any_Moment_CVPR_2026_paper.html}
}

@misc{resight_remembr,
  title = {{ReMEmbR: Building and Reasoning Over Long-Horizon Spatio-Temporal Memory for Robot Navigation}},
  author = {Anwar, Abrar and Welsh, John and Biswas, Joydeep and Pouya, Soha and Chang, Yan},
  year = {2024},
  eprint = {2409.13682},
  archivePrefix = {arXiv},
  url = {https://arxiv.org/abs/2409.13682}
}

@misc{resight_raven,
  title = {{RAVEN: Long-Horizon Reasoning \& Navigation with a Visuo-Spatio-Temporal Memory}},
  author = {Hu, Yixun and Zheng, Zhicheng and Zha, Lihan and Xing, Chunwei and Singh, Rajdeep and Hossain, Omar and Loquercio, Antonio and Shah, Dhruv},
  year = {2026},
  eprint = {2606.25206},
  archivePrefix = {arXiv},
  url = {https://arxiv.org/abs/2606.25206}
}

@misc{resight_mom,
  title = {{Memory Over Maps: 3D Object Localization Without Reconstruction}},
  author = {Zhou, Rui and Yap, Xander and Cao, Jianwen and Lau, Allison and Sun, Boyang and Pollefeys, Marc},
  year = {2026},
  eprint = {2603.20530},
  archivePrefix = {arXiv},
  url = {https://arxiv.org/abs/2603.20530}
}

@misc{resight_worth,
  title = {{Worth Remembering: Surprise-Gated Robot Episodic Memory}},
  author = {Gorlo, Nicolas and Wise, Derek K. and Speranzon, Alberto and Carlone, Luca},
  year = {2026},
  eprint = {2606.03787},
  archivePrefix = {arXiv},
  url = {https://arxiv.org/abs/2606.03787}
}

@inproceedings{resight_cap,
  title = {{Code as Policies: Language Model Programs for Embodied Control}},
  author = {Liang, Jacky and Huang, Wenlong and Xia, Fei and Xu, Peng and Hausman, Karol and Ichter, Brian and Florence, Pete and Zeng, Andy},
  booktitle = {Proceedings of the IEEE International Conference on Robotics and Automation (ICRA)},
  year = {2023},
  pages = {9493--9500},
  url = {https://doi.org/10.1109/ICRA48891.2023.10160591}
}

@misc{resight_viper,
  title = {{ViperGPT: Visual Inference via Python Execution for Reasoning}},
  author = {Surís, Dídac and Menon, Sachit and Vondrick, Carl},
  year = {2023},
  eprint = {2303.08128},
  archivePrefix = {arXiv},
  url = {https://arxiv.org/abs/2303.08128}
}

@inproceedings{resight_latm,
  title = {{Large Language Models as Tool Makers}},
  author = {Cai, Tianle and Wang, Xuezhi and Ma, Tengyu and Chen, Xinyun and Zhou, Denny},
  booktitle = {International Conference on Learning Representations (ICLR)},
  year = {2024},
  pages = {54067--54089},
  url = {https://openreview.net/forum?id=qV83K9d5WB}
}

@misc{resight_voyager,
  title = {{Voyager: An Open-Ended Embodied Agent with Large Language Models}},
  author = {Wang, Guanzhi and Xie, Yuqi and Jiang, Yunfan and Mandlekar, Ajay and Xiao, Chaowei and Zhu, Yuke and Fan, Linxi and Anandkumar, Anima},
  year = {2023},
  eprint = {2305.16291},
  archivePrefix = {arXiv},
  url = {https://arxiv.org/abs/2305.16291}
}

@misc{resight_aspire,
  title = {{ASPIRE: Agentic /Skills Discovery for Robotics}},
  author = {Lu, Runyu and Wu, Yubo and Kou, Ethan and Fu, Letian and Xiao, Wenli and Mandlekar, Ajay and Xu, Yinzhen and Shi, Guanya and Goldberg, Ken and Chen, Ang and Chowdhury, Mosharaf and Zhu, Yuke and Fan, Linxi "Jim" and Wang, Guanzhi},
  year = {2026},
  eprint = {2607.00272},
  archivePrefix = {arXiv},
  url = {https://arxiv.org/abs/2607.00272}
}

@misc{resight_metaharness,
  title = {{Meta-Harness: End-to-End Optimization of Model Harnesses}},
  author = {Lee, Yoonho and Nair, Roshen and Zhang, Qizheng and Lee, Kangwook and Khattab, Omar and Finn, Chelsea},
  year = {2026},
  eprint = {2603.28052},
  archivePrefix = {arXiv},
  url = {https://arxiv.org/abs/2603.28052}
}

@inproceedings{resight_clipfields,
  title = {{CLIP-Fields: Weakly Supervised Semantic Fields for Robotic Memory}},
  author = {Shafiullah, Nur Muhammad Mahi and Paxton, Chris and Pinto, Lerrel and Chintala, Soumith and Szlam, Arthur},
  booktitle = {Proceedings of Robotics: Science and Systems (RSS)},
  year = {2023},
  url = {https://www.roboticsproceedings.org/rss19/p074.html}
}

@misc{resight_openmask3d,
  title = {{OpenMask3D: Open-Vocabulary 3D Instance Segmentation}},
  author = {Takmaz, Ayça and Fedele, Elisabetta and Sumner, Robert W. and Pollefeys, Marc and Tombari, Federico and Engelmann, Francis},
  year = {2023},
  eprint = {2306.13631},
  archivePrefix = {arXiv},
  url = {https://arxiv.org/abs/2306.13631}
}

@misc{resight_hovsg,
  title = {{Hierarchical Open-Vocabulary 3D Scene Graphs for Language-Grounded Robot Navigation}},
  author = {Werby, Abdelrhman and Huang, Chenguang and Büchner, Martin and Valada, Abhinav and Burgard, Wolfram},
  year = {2024},
  eprint = {2403.17846},
  archivePrefix = {arXiv},
  url = {https://arxiv.org/abs/2403.17846}
}

@inproceedings{resight_dynamem,
  title = {{DynaMem: Online Dynamic Spatio-Semantic Memory for Open World Mobile Manipulation}},
  author = {Liu, Peiqi and Guo, Zhanqiu and Warke, Mohit and Chintala, Soumith and Paxton, Chris and Shafiullah, Nur Muhammad Mahi and Pinto, Lerrel},
  booktitle = {Proceedings of the IEEE International Conference on Robotics and Automation (ICRA)},
  year = {2025},
  pages = {13346--13355},
  url = {https://doi.org/10.1109/ICRA55743.2025.11127619}
}

@misc{resight_khronos,
  title = {{Khronos: A Unified Approach for Spatio-Temporal Metric-Semantic SLAM in Dynamic Environments}},
  author = {Schmid, Lukas and Abate, Marcus and Chang, Yun and Carlone, Luca},
  year = {2024},
  eprint = {2402.13817},
  archivePrefix = {arXiv},
  url = {https://arxiv.org/abs/2402.13817}
}

@misc{resight_clio,
  title = {{Clio: Real-time Task-Driven Open-Set 3D Scene Graphs}},
  author = {Maggio, Dominic and Chang, Yun and Hughes, Nathan and Trang, Matthew and Griffith, Dan and Dougherty, Carlyn and Cristofalo, Eric and Schmid, Lukas and Carlone, Luca},
  year = {2024},
  eprint = {2404.13696},
  archivePrefix = {arXiv},
  url = {https://arxiv.org/abs/2404.13696}
}

@misc{resight_embodiedrag,
  title = {{Embodied-RAG: General Non-parametric Embodied Memory for Retrieval and Generation}},
  author = {Xie, Quanting and Min, So Yeon and Ji, Pengliang and Yang, Yue and Zhang, Tianyi and Xu, Kedi and Bajaj, Aarav and Salakhutdinov, Ruslan and Johnson-Roberson, Matthew and Bisk, Yonatan},
  year = {2024},
  eprint = {2409.18313},
  archivePrefix = {arXiv},
  url = {https://arxiv.org/abs/2409.18313}
}

@misc{resight_threedmem,
  title = {{3D-Mem: 3D Scene Memory for Embodied Exploration and Reasoning}},
  author = {Yang, Yuncong and Yang, Han and Zhou, Jiachen and Chen, Peihao and Zhang, Hongxin and Du, Yilun and Gan, Chuang},
  year = {2024},
  eprint = {2411.17735},
  archivePrefix = {arXiv},
  url = {https://arxiv.org/abs/2411.17735}
}

@inproceedings{resight_keysg,
  title = {{KeySG: Hierarchical Keyframe-Based 3D Scene Graphs}},
  author = {Werby, Abdelrhman and Rotondi, Dennis and Scaparro, Fabio and Arras, Kai O.},
  booktitle = {Proceedings of the IEEE International Conference on Robotics and Automation (ICRA)},
  year = {2026},
  url = {https://keysg-lab.github.io/}
}

@misc{resight_metamemory,
  title = {{Meta-Memory: Retrieving and Integrating Semantic-Spatial Memories for Robot Spatial Reasoning}},
  author = {Mao, Yufan and Ye, Hanjing and Dong, Wenlong and Zhang, Chengjie and Zhang, Hong},
  year = {2025},
  eprint = {2509.20754},
  archivePrefix = {arXiv},
  url = {https://arxiv.org/abs/2509.20754}
}

@misc{resight_threedllmmem,
  title = {{3DLLM-Mem: Long-Term Spatial-Temporal Memory for Embodied 3D Large Language Model}},
  author = {Hu, Wenbo and Hong, Yining and Wang, Yanjun and Gao, Leison and Wei, Zibu and Yao, Xingcheng and Peng, Nanyun and Bitton, Yonatan and Szpektor, Idan and Chang, Kai-Wei},
  year = {2025},
  eprint = {2505.22657},
  archivePrefix = {arXiv},
  url = {https://arxiv.org/abs/2505.22657}
}

@inproceedings{resight_openscene,
  title = {{OpenScene: 3D Scene Understanding with Open Vocabularies}},
  author = {Peng, Songyou and Genova, Kyle and Jiang, Chiyu "Max" and Tagliasacchi, Andrea and Pollefeys, Marc and Funkhouser, Thomas},
  booktitle = {Proceedings of the IEEE/CVF Conference on Computer Vision and Pattern Recognition (CVPR)},
  year = {2023},
  pages = {815--824},
  url = {https://doi.org/10.1109/CVPR52729.2023.00085}
}

@misc{resight_conceptfusion,
  title = {{ConceptFusion: Open-set Multimodal 3D Mapping}},
  author = {Jatavallabhula, Krishna Murthy and Kuwajerwala, Alihusein and Gu, Qiao and Omama, Mohd and Chen, Tao and Maalouf, Alaa and Li, Shuang and Iyer, Ganesh and Saryazdi, Soroush and Keetha, Nikhil and Tewari, Ayush and Tenenbaum, Joshua B. and de Melo, Celso Miguel and Krishna, Madhava and Paull, Liam and Shkurti, Florian and Torralba, Antonio},
  year = {2023},
  eprint = {2302.07241},
  archivePrefix = {arXiv},
  url = {https://arxiv.org/abs/2302.07241}
}

@misc{resight_m3,
  title = {{M3: 3D-Spatial MultiModal Memory}},
  author = {Zou, Xueyan and Song, Yuchen and Qiu, Ri-Zhao and Peng, Xuanbin and Ye, Jianglong and Liu, Sifei and Wang, Xiaolong},
  year = {2025},
  eprint = {2503.16413},
  archivePrefix = {arXiv},
  url = {https://arxiv.org/abs/2503.16413}
}

@misc{resight_foundit,
  title = {{FOUND-IT: Foundation-model-first Task-driven 3D Scene Graphs with Granularity on Demand}},
  author = {Maggio, Dominic and Gorlo, Nicolas and Hauser, Kris and Carlone, Luca},
  year = {2026},
  eprint = {2605.25371},
  archivePrefix = {arXiv},
  url = {https://arxiv.org/abs/2605.25371}
}

@misc{resight_videoagent,
  title = {{VideoAgent: A Memory-augmented Multimodal Agent for Video Understanding}},
  author = {Fan, Yue and Ma, Xiaojian and Wu, Rujie and Du, Yuntao and Li, Jiaqi and Gao, Zhi and Li, Qing},
  year = {2024},
  eprint = {2403.11481},
  archivePrefix = {arXiv},
  url = {https://arxiv.org/abs/2403.11481}
}

@misc{resight_mobilityvla,
  title = {{Mobility VLA: Multimodal Instruction Navigation with Long-Context VLMs and Topological Graphs}},
  author = {Chiang, Hao-Tien Lewis and Xu, Zhuo and Fu, Zipeng and Jacob, Mithun George and Zhang, Tingnan and Lee, Tsang-Wei Edward and Yu, Wenhao and Schenck, Connor and Rendleman, David and Shah, Dhruv and Xia, Fei and Hsu, Jasmine and Hoech, Jonathan and Florence, Pete and Kirmani, Sean and Singh, Sumeet and Sindhwani, Vikas and Parada, Carolina and Finn, Chelsea and Xu, Peng and Levine, Sergey and Tan, Jie},
  year = {2024},
  eprint = {2407.07775},
  archivePrefix = {arXiv},
  url = {https://arxiv.org/abs/2407.07775}
}

@inproceedings{resight_gepa,
  title = {{GEPA: Reflective Prompt Evolution Can Outperform Reinforcement Learning}},
  author = {Agrawal, Lakshya A and Tan, Shangyin and Soylu, Dilara and Ziems, Noah and Khare, Rishi and Opsahl-Ong, Krista and Singhvi, Arnav and Shandilya, Herumb and Ryan, Michael J and Jiang, Meng and Potts, Christopher and Sen, Koushik and Dimakis, Alex and Stoica, Ion and Klein, Dan and Zaharia, Matei and Khattab, Omar},
  booktitle = {International Conference on Learning Representations (ICLR)},
  year = {2026},
  pages = {8479--8565},
  url = {https://openreview.net/forum?id=RQm2KQTM5r}
}

@misc{resight_ace,
  title = {{Agentic Context Engineering: Evolving Contexts for Self-Improving Language Models}},
  author = {Zhang, Qizheng and Hu, Changran and Upasani, Shubhangi and Ma, Boyuan and Hong, Fenglu and Kamanuru, Vamsidhar and Rainton, Jay and Wu, Chen and Ji, Mengmeng and Li, Hanchen and Thakker, Urmish and Zou, James and Olukotun, Kunle},
  year = {2025},
  eprint = {2510.04618},
  archivePrefix = {arXiv},
  url = {https://arxiv.org/abs/2510.04618}
}

@inproceedings{resight_adas,
  title = {{Automated Design of Agentic Systems}},
  author = {Hu, Shengran and Lu, Cong and Clune, Jeff},
  booktitle = {International Conference on Learning Representations (ICLR)},
  year = {2025},
  pages = {21344--21377},
  url = {https://openreview.net/forum?id=t9U3LW7JVX}
}

@misc{resight_sica,
  title = {{A Self-Improving Coding Agent}},
  author = {Robeyns, Maxime and Szummer, Martin and Aitchison, Laurence},
  year = {2025},
  eprint = {2504.15228},
  archivePrefix = {arXiv},
  url = {https://arxiv.org/abs/2504.15228}
}

@misc{resight_dgm,
  title = {{Darwin Godel Machine: Open-Ended Evolution of Self-Improving Agents}},
  author = {Zhang, Jenny and Hu, Shengran and Lu, Cong and Lange, Robert and Clune, Jeff},
  year = {2025},
  eprint = {2505.22954},
  archivePrefix = {arXiv},
  url = {https://arxiv.org/abs/2505.22954}
}

@misc{resight_autoharness,
  title = {{AutoHarness: improving LLM agents by automatically synthesizing a code harness}},
  author = {Lou, Xinghua and Lázaro-Gredilla, Miguel and Dedieu, Antoine and Wendelken, Carter and Lehrach, Wolfgang and Murphy, Kevin P.},
  year = {2026},
  eprint = {2603.03329},
  archivePrefix = {arXiv},
  url = {https://arxiv.org/abs/2603.03329}
}

@misc{resight_ahe,
  title = {{Agentic Harness Engineering: Observability-Driven Automatic Evolution of Coding-Agent Harnesses}},
  author = {Lin, Jiahang and Liu, Shichun and Pan, Chengjun and Lin, Lizhi and Dou, Shihan and Xi, Zhiheng and Huang, Xuanjing and Yan, Hang and Han, Zhenhua and Gui, Tao and Jiang, Yu-Gang},
  year = {2026},
  eprint = {2604.25850},
  archivePrefix = {arXiv},
  url = {https://arxiv.org/abs/2604.25850}
}

@misc{resight_selfharness,
  title = {{Self-Harness: Harnesses That Improve Themselves}},
  author = {Zhang, Hangfan and Zhang, Shao and Li, Kangcong and Zhang, Chen and Chen, Yang and Zhang, Yiqun and Bai, Lei and Hu, Shuyue},
  year = {2026},
  eprint = {2606.09498},
  archivePrefix = {arXiv},
  url = {https://arxiv.org/abs/2606.09498}
}

@misc{resight_harnessx,
  title = {{HarnessX: A Composable, Adaptive, and Evolvable Agent Harness Foundry}},
  author = {Chen, Tingyang and Lu, Shuo and Zhao, Kang and Meng, Weicheng and Teng, Hanlin and Li, Tianhao and Li, Chao and Liu, Xule and Liang, Jian and Zhang, Zhizhong and Xie, Yuan and Qu, Heng and Shao, Kun and Luan, Jian},
  year = {2026},
  eprint = {2606.14249},
  archivePrefix = {arXiv},
  url = {https://arxiv.org/abs/2606.14249}
}

@misc{resight_harnesscompass,
  title = {{HarnessCompass: Guiding Automatic Harness Evolution toward Generalizable and Effective Agent Harnesses}},
  author = {Zhang, Luan and Zhou, Ruochen and Song, Dandan and Chen, Zhengyu and Tian, Yuhang and Yang, Jun and Ma, Huipeng and Li, Chenhao and Feng, Guangyuan and Li, Xudong and Jin, Yizhou and Xu, Yan},
  year = {2026},
  eprint = {2608.01918},
  archivePrefix = {arXiv},
  url = {https://arxiv.org/abs/2608.01918}
}

@misc{resight_demoevolve,
  title = {{DemoEvolve: Overcoming Sparse Feedback in Agentic Harness Evolution with Demonstrations}},
  author = {Che, Lirong and Yang, Yuzhe and Lin, Peiwen and Wang, Chuang and Wang, Xueqian and Su, Jian},
  year = {2026},
  eprint = {2605.24539},
  archivePrefix = {arXiv},
  url = {https://arxiv.org/abs/2605.24539}
}

@misc{resight_memevolve,
  title = {{MemEvolve: Meta-Evolution of Agent Memory Systems}},
  author = {Zhang, Guibin and Ren, Haotian and Zhan, Chong and Zhou, Zhenhong and Wang, Junhao and Zhu, He and Zhou, Wangchunshu and Yan, Shuicheng},
  year = {2025},
  eprint = {2512.18746},
  archivePrefix = {arXiv},
  url = {https://arxiv.org/abs/2512.18746}
}

@misc{resight_lagmemo,
  title = {{LagMemo: Language 3D Gaussian Splatting Memory for Multi-modal Open-vocabulary Multi-goal Visual Navigation}},
  author = {Zhou, Haotian and Wang, Xiaole and Li, He and Qi, Zhuo and Yin, Jinrun and Kong, Haiyu and Xu, Jianghuan and Zhao, Huijing},
  year = {2025},
  eprint = {2510.24118},
  archivePrefix = {arXiv},
  primaryClass = {cs.RO},
  url = {https://arxiv.org/abs/2510.24118}
}

@inproceedings{resight_openfungraph,
  title = {{Open-Vocabulary Functional 3D Scene Graphs for Real-World Indoor Spaces}},
  author = {Zhang, Chenyangguang and Delitzas, Alexandros and Wang, Fangjinhua and Zhang, Ruida and Ji, Xiangyang and Pollefeys, Marc and Engelmann, Francis},
  booktitle = {Proceedings of the IEEE/CVF Conference on Computer Vision and Pattern Recognition (CVPR)},
  year = {2025},
  pages = {19401--19413},
  url = {https://doi.org/10.1109/CVPR52734.2025.01807}
}
\bibliographystyle{iclr2027_conference}

\clearpage

\appendix

\raggedbottom
\section{Base Memory and Tools}
\label{app:baselines}

\subsection{Base Tool Interface}
\label{app:interface}
Table~\ref{tab:app_tools} gives the arguments and function of the base tools used on OC-NaVQA and FindingDory. Brackets mark optional arguments. Retrieval limits and online rounds are given in Appendix~\ref{app:ocnavqa}.

\begin{table}[ht]
\centering
\caption{Base execution interface for OC-NaVQA and FindingDory. Retrieval tools return text summaries; position recall requires camera poses.}
\label{tab:app_tools}

\setlength{\tabcolsep}{4pt}
\renewcommand{\arraystretch}{1.05}
\begin{tabular}{@{}p{0.30\linewidth}p{0.40\linewidth}p{0.22\linewidth}@{}}
\toprule
\textbf{Tool} & \textbf{Arguments} & \textbf{Function} \\
\midrule
\texttt{semantic\_search} & \texttt{query}, [\texttt{top\_k}] & RRF semantic search \\
\texttt{recall\_around\_time} & \texttt{clock\_time}, [\texttt{limit}] & Time-neighbor recall \\
\texttt{recall\_near\_position} & \texttt{position} $[x,y,z]$, [\texttt{limit}] & Pose-neighbor recall \\
\texttt{inspect\_frames} & \texttt{frame\_ids}, [\texttt{focus}] & Show original images \\
\texttt{submit\_answer} & \texttt{type}, \texttt{reasoning}, \texttt{source\_frame\_ids}, type-specific answer field & Submit answer \\
\bottomrule
\end{tabular}
\end{table}

Position retrieval returns the camera location of an observation. Estimating an object's location within that image requires geometry; recovering an interaction sequence requires reasoning across frames. EDM adds tools for these operations.

\subsection{Retrieval Fusion and Progressive Disclosure}
\label{app:retrieval-fusion}
Content retrieval combines a SigLIP2 visual embedding with e5 embeddings of the scene description and per-frame object text. Each channel ranks candidates independently. Reciprocal rank fusion then scores frame $i$ for query $q$ by
\begin{equation}
\operatorname{score}(i,q)=\sum_{c\in\mathcal C}\frac{w_c}{\kappa+\operatorname{rank}_c(i,q)}.
\end{equation}
A time interval can filter candidates. Where poses exist, nearby viewpoints can be grouped for retrieval while each frame remains individually accessible.

The first return is compact text: frame id, description, object fields, and available temporal and spatial metadata, cropped and formatted from stored records. The agent opens the original images only when it needs to verify text, object state, or visual detail. This separates candidate screening from image inspection; measured online usage appears in Appendix~\ref{app:budgets}.

\subsection{Keyframe Fields and an Annotation Example}
\label{app:keyframe-fields}
Each retained keyframe links an image through \texttt{frame\_id} to its capture time and, when available, camera pose and depth. These metadata come from acquisition and processing. A vision-language annotator reads the image without seeing the later questions and generates semantic text.

The following excerpt is from frame 0 of OC-NaVQA sequence 0, annotated by Qwen3.8-27B. It shows the scene description and the first two objects; other fields and objects are omitted:
\noindent\begin{minipage}{\linewidth}
\begin{lstlisting}[
  basicstyle=\ttfamily,
  columns=fullflexible,
  keepspaces=true,
  showstringspaces=false,
  breaklines=true,
  breakatwhitespace=true,
  breakindent=1em,
  backgroundcolor=\color{black!3},
  frame=single,
  rulecolor=\color{black!20},
  framesep=6pt,
  xleftmargin=6pt,
  xrightmargin=6pt,
  aboveskip=\baselineskip,
  belowskip=\baselineskip
]
{
  "global_description": "An outdoor industrial alley or service walkway flanked by brick buildings, containing large green utility structures, a gray electrical or mechanical housing unit, and a paved concrete floor.",
  "objects": [
    {
      "label": "green utility enclosure",
      "description": "A large, long, dark green rectangular structure, likely a utility tunnel cover or storage container, sitting on the left side of the alley."
    },
    {
      "label": "gray utility housing",
      "description": "A light gray, box-like metal cabinet or small structure with vents and a door, located at the end of the concrete path."
    }
  ]
}
\end{lstlisting}
\end{minipage}

The scene description captures layout; object labels and descriptions provide appearance and local context. Both text types are encoded separately and linked to the image through \texttt{frame\_id}. Later memory builders can reprocess retained images for OCR, interactions, or geometry without collecting the observations again. The annotator replacement experiment is in Appendix~\ref{app:annotator}.

\subsection{Two Ways to Access the Same Memory}
\label{app:access}
The execution agent uses registered tools and mounted skills within an online round limit. The coding agent reads observations and processing programs in a workspace, inspects images, writes code, and prepares models or dependencies during exploration. The capability library holds tools, skills, and memory builders. EDM adds reusable artifacts from exploration to that library for subsequent execution.

\section{Explore--Distill--Mount Implementation}
\label{app:edm-implementation}

\subsection{Diagnosing Capability Gaps from Execution}
\label{app:edm-diagnosis}
After exploration, distillation inventories the host's tools, memory fields, and execution traces in a capability ledger. Each operation used during exploration is classified as directly supported, composable from existing operations, partially supported, or missing. A script flags queries whose traces show a difficulty, such as a forced final answer, repeated identical calls, or an answer given without inspecting frames, and queries on which the executor's answer disagrees with an explorer's answer. A gap map groups the flagged queries by cause and relates each gap to the operations and exploration programs that address it.

Diagnosis uses execution traces and answer-disagreement flags to identify candidate gaps, without task reference answers or evaluation scores.

\subsection{Exploration, Distillation, and Mounting}
\label{app:edm-stages}
In \emph{Explore}, one coding-agent session per memory answers all queries posed on that memory. The agent inspects retained observations, writes programs, and delivers an answer to every query together with its programs, notes, and supporting frames. A program may answer a query directly or first construct a field for the whole memory, such as an OCR index or a pose trajectory.

In \emph{Distill}, the agent compares these operations with the capability ledger. Missing or partially supported operations become tools or memory builders, each linked to a diagnosed gap and its evidence, with the exploration programs as starting implementations. An optional replay round mounts the tools, lets the executor answer queries with them, and revises the tools from the resulting traces. A skill then describes when and how the executor should use the tools.

In \emph{Mount}, distilled tools are registered, the skill, if any, enters the executor's context, and builders expose their fields through query tools. Whole-memory processing and model initialization can run before queries; query-dependent computation runs when a tool is called. The execution model and online budget remain the same before and after mounting. For example, an OCR builder stores text with its source \texttt{frame\_id}; a text-query tool then returns matching frames, which the executor may inspect. The builder can be rerun over a new patrol memory.

\subsection{EDM Procedure}
\label{app:edm-algorithm}
Algorithm~\ref{alg:edm} lists one EDM round. \textsc{Run}, \textsc{Flag}, and \textsc{Mount} are scripts; each other step is a coding-agent session whose prompt is given in Appendix~\ref{app:edm-prompts}. $B$ is the online budget used on both sides.

\begin{algorithm}[t]
\caption{One EDM round}
\label{alg:edm}
\begin{algorithmic}[1]
\Require execution configuration $H$, memories $\{\mathcal M_e\}$ with queries $\{\mathcal Q_e\}$, online budget $B$
\Ensure mounted configuration $H'$ under the same budget $B$
\State $\mathcal T\gets\textsc{Run}(H,\bigcup_e\mathcal Q_e,B)$ \Comment{traces, without correctness labels}
\For{each memory $e$, in parallel}
  \State $\mathcal E_e\gets\textsc{Explore}(\mathcal M_e,\mathcal Q_e)$ \Comment{answers, programs, notes}
\EndFor
\State $F\gets\textsc{Flag}(\mathcal T,\mathcal E)$ \Comment{behavior flags; answers that differ from the explorers'}
\State $L,G\gets\textsc{Diagnose}(H,\mathcal T,F,\mathcal E)$ \Comment{capability ledger, gap map}
\State $\mathcal A\gets\textsc{Build}(H,\mathcal T,L,G,\mathcal E)$ \Comment{tools and builders for the gaps}
\For{$r=1,\dots,R$} \Comment{optional replay; $R=0$ skips refinement}
  \State $\mathcal T_r\gets\textsc{Run}(\textsc{Mount}(H,\mathcal A),\mathcal Q_r,B)$;\ \ $\mathcal A\gets\textsc{Refine}(\mathcal A,\mathcal T_r)$
\EndFor
\State $K\gets\textsc{Skill}(\mathcal A,\mathcal E,\mathcal T)$
\State \Return $\textsc{Mount}(H,\mathcal A,K)$
\end{algorithmic}
\end{algorithm}

\subsection{EDM Prompts}
\label{app:edm-prompts}
\newcommand{\edmph}[1]{\texttt{\{#1\}}}
\newcommand{\edmprompt}[1]{\leavevmode\par\nopagebreak\noindent\hspace*{6pt}{\setlength{\fboxsep}{6pt}\fcolorbox{black!20}{black!3}{\parbox{\dimexpr\linewidth-12pt-2\fboxsep-2\fboxrule\relax}{\setlength{\parskip}{3pt} #1}}}\par\medskip}
\textsc{Explore}, \textsc{Diagnose}, \textsc{Build}, \textsc{Refine}, and \textsc{Skill} in Algorithm~\ref{alg:edm} are coding-agent sessions. Below we give the instructions that define each session, with its inputs, deliverables, and constraints; operational details such as file layout, output schemas, and session management are omitted. Text in braces stands for inputs that vary across benchmarks, such as the memory, the executor's tools, and its online budget (Appendix~\ref{app:setup}).

\paragraph{Explore (one session per memory).}
\edmprompt{You are a coding agent with full access to this workspace. It holds \edmph{memory}: \edmph{memory description}, and \edmph{query set}: the queries posed on this memory, without reference answers. Work on all the queries in this session.

1.~Answer every query yourself, by any method: inspect images, write and run programs, compute statistics, or use models. Support each answer with frames you have checked.

2.~Keep each operation you used as a program that can be run again.

Deliverables: an answer to every query, with the frames that support it; programs for these operations, each opening with what it computes and how to call it; notes on the facts you measured, the query types each operation serves, and the pitfalls you encountered; and, optionally, operations already packaged as tools or builders that follow \edmph{interface}.

Do not read, search for, or infer reference answers or evaluation results.}

\paragraph{Diagnose.}
\edmprompt{The execution agent has \edmph{executor tools} and \edmph{online budget}. \edmph{execution traces} record its behavior on the task pool. \edmph{flags} mark the queries to examine: queries whose behavior shows a difficulty, and queries where its answer differs from an explorer's answer. A flag is a reason for doubt, not a judgment of correctness, and no reference answers are involved. \edmph{exploration records} contain the explorers' notes and programs. Do not write code in this step.

1.~Capability ledger. List each operation the explorers used, merging repeats across recordings. Classify it against the execution agent as direct (a base tool returns the same evidence), composable (a combination of base tools returns it), partial (a related base tool lacks an essential affordance, which you name), or missing. Cite the tool description or the execution traces that justify each class.

2.~Gap map. Group the flagged queries by the cause behind their flags. For each gap, list the affected queries and their \edmph{query categories}, the flags that expose it, the operations that would resolve it, and the exploration programs that implement them. Assign a remedy: none if every operation it needs is direct or composable, a builder if an operation needs one pass over the whole memory, and a tool otherwise.

Deliverables: the capability ledger and the gap map.}

\paragraph{Build.}
\edmprompt{Build only what the gap map calls for. You have the capability ledger and the gap map from the diagnosis, which you take as given; the execution agent's \edmph{executor tools} and \edmph{execution traces}; \edmph{exploration records}, whose programs are the implementations to start from; and recordings to test your tools on.

1.~For each gap whose remedy is a tool or a builder, build it from the exploration programs the gap lists, reusing any other exploration program or packaged tool that helps. Implement a tool for computation at query time and a builder for computation over the whole memory; tools read the builders' outputs.

2.~For a partial operation, extend the base tool's evidence with the missing affordance instead of building a parallel tool. Never rebuild a direct or composable operation.

3.~Tools follow \edmph{interface}. They return evidence, such as frame identifiers, times, tables, or rendered images, and never answers. Replies stay within \edmph{reply and time limits}. Thresholds are measured on the memory at hand, with no constants specific to one recording.

4.~Each tool's description states when the execution agent should call it and how it complements the base tools.

Run every tool on the test recordings before submitting. Deliverables: the tools and builders, and a mapping from each tool to the gaps it closes.}

\paragraph{Refine (optional).}
\edmprompt{The distilled tools were mounted, and the execution agent answered \edmph{replay queries} with them; \edmph{replay traces} record these runs, without correctness labels. Examine how the tools' replies shaped its behavior: calls spent completing a reply, fields it misread, candidates it could not tell apart, and answers that differ across repeated trials. Revise the tools at the level of method, without special cases for particular queries or recordings, and without adding tools for operations the ledger classes as direct or composable.

Deliverables: the revised tools, and a note on what changed and which queries showed each problem.}

\paragraph{Skill.}
\edmprompt{The execution agent now has the distilled tools \edmph{distilled tools} in addition to \edmph{executor tools}, under \edmph{online budget}. Write a skill of at most \edmph{N} lines that will be appended to its system prompt. State which tool to call first for which query type, in what order, what counts as sufficient evidence, which heuristics mislead, and how to spend the budget. Base each line on the explorers' notes or the execution traces, and end it with its source. Do not state or hint at any query's answer, and do not mention frame numbers, times, recording or query identifiers, or object and room names; write only what holds across recordings.

Deliverable: the skill, one instruction per line, each line ending with its source.}

\section{Experimental Settings and Comparison Protocol}
\label{app:setup}

\subsection{Comparison Principles and Reproduced Configurations}
\label{app:baseline-configs}
Within a host, the before-and-after comparison fixes the execution model, base interface, decision-round cap, input, and scoring rule. Across hosts we retain each baseline's memory representation and agent loop and report their measured rounds, tokens, and image access. The coding-agent exploration reference solves an episode jointly with a separate workspace and compute budget.

For GOAT-core, the visual-only MoM reproduction uses SigLIP2 retrieval with the common segmentation, depth-backprojection, and multi-view localization backend. ReSight's retrieval-only rows change the retrieval representation and interface but do not invoke a query-time language model to choose candidates. The DAAAM reproduction uses its own scene graph and agent loop on OC-NaVQA, receives no original images online, and has no additional decision-round cap. RAVEN retains its benchmark-specific loop and image-bearing retrieval; on FindingDory, its temporal return size is reduced from 200 to 30 to limit single-request image input.

\subsection{GOAT-core}
\label{app:goat-models}
GOAT-core has 480 image, object, and language queries across four scenes and 1,606 retained keyframes. A predicted location succeeds when it lies within 1.5\,m of the target on the horizontal XZ plane. Retrieval-only and agent configurations share the localization backend. SR@1, SR@3, and SR@5 use prefixes of the same ranked candidate list; the main table reports SR@5.

The image encoder is SigLIP2-SO400M with patch size 14 at $384\times384$ resolution; e5 encodes text fields. Each enabled retrieval channel returns 40 candidates, combined with equal-weight RRF and $\kappa=60$. When poses exist, viewpoint grouping uses 0.75\,m and $25^\circ$ thresholds. Retrieval-only returns at most five candidate views. The GPT-5.6-sol agent runs at most three rounds; semantic search returns eight entries by default and at most twelve, \texttt{inspect\_frames} opens at most four images, and \texttt{submit\_result} sends the candidates. The writing-time annotator comparison uses GPT-5.6-sol and Qwen2.5-VL-7B.

\subsection{OC-NaVQA and FindingDory}
\label{app:ocnavqa}
\label{app:findingdory}
Both benchmarks use GPT-5.6-sol in the same ReSight execution loop. Semantic retrieval combines visual, scene-description, and object-text channels, each with 40 candidates; RRF uses equal weights and $\kappa=60$. Base retrieval returns text, while \texttt{inspect\_frames} opens at most four low-detail images per call. FindingDory has no position recall because its data lack camera poses.

\begin{table}[ht]
\centering
\setlength{\tabcolsep}{3pt}
\renewcommand{\arraystretch}{1.12}
\begin{tabular}{p{0.23\linewidth}p{0.31\linewidth}p{0.27\linewidth}c}
\toprule
Host / benchmark & Semantic / temporal return & Position return & Rounds \\
\midrule
ReSight / OC & 12 text / 8 default, 12 max & 8 default, 12 max text & 6 \\
ReSight / FD & 12 text / 8 default, 12 max & None & 15 \\
RAVEN / OC & 5 each, with images & 5 with images & 3 \\
RAVEN / FD & 30 each, with images & None & 15 \\
\bottomrule
\end{tabular}
\caption{Base retrieval limits per call and decision-round caps. EDM keeps each host's limits unchanged. Added tools can return event summaries, frame ids, or rendered panels.}
\label{tab:app_fd_retrieval}
\end{table}

On FindingDory both hosts cap each request at 30 images and manage image history according to their own loop. ReSight's rendered panels use a separate image configuration. One ReSight round is one model request and may emit several tool calls. OC-NaVQA allows six decision rounds and an additional finalization request if no answer was submitted; RAVEN allows three tool-calling decisions followed by finalization and answer generation. Appendix~\ref{app:budgets} counts all online model requests.

OC-NaVQA contains 210 questions in seven sequences and 3,973 keyframes. Position ground truth is an object coordinate; ReSight without EDM and the RAVEN adaptation return the selected observation's camera position, whereas the mounted position tool estimates an object coordinate. We report mean position error and the share within 10\,m. Temporal MAE pools 30 timestamp and nine duration questions, in minutes. Text answers follow the common three-vote judge protocol.

Gains concentrate on localization, a major gap identified by the ledger; localization improves with both annotators (Table~\ref{tab:app_annotator_oc}, Appendix~\ref{app:annotator}). With Sol annotations, mounting EDM corrects two of the 67 binary answers but introduces five errors (exact McNemar $p=0.45$); binary accuracy remains unchanged with Qwen annotations. Text accuracy is unchanged in both settings, while temporal error decreases with Sol and increases with Qwen.

FindingDory evaluates 596 questions in episodes 91--100: 297 spatial, 210 temporal, and 89 multi-goal. A frame is retained when its SigLIP2 feature has cosine similarity below 0.90 to the preceding retained frame, yielding 4,660 keyframes with scene and object annotations. Without poses, each frame forms its own viewpoint group. The official high-level success rule scores each query; Overall is the micro-average across 596 questions. Submitted frame ids are converted to memory time through the episode mapping, with ordered or unordered evidence as requested and an empty list for an absent target.

The two FindingDory hosts each perform Distill and Mount against their own capability ledger, reusing exploration material from episodes 91--100. For the main results on these episodes, diagnosis and evaluation use the same public task pool and memory; the EDM procedure is described in Appendix~\ref{app:edm-algorithm}.

\paragraph{Frozen-capability evaluation on episodes 81--90.}
We evaluate 597 queries from episodes 81--90, excluded from exploration and distillation, using the frozen tools and skill developed on episodes 91--100. The same memory-construction pipeline yields 4,016 keyframes. Both configurations use GPT-5.6-sol through the same API backend, with 15 decision rounds and at most 30 images per request. We use the official high-level success rule and micro-average over all 597 queries. Tokens per query include model input and output; results appear in Table~\ref{tab:fd-heldout}.

\paragraph{Distillation without exploration.}
\label{app:fd-wo-explore}
This configuration repeats Distill and Mount for episodes 91--100 with the exploration stage removed. The coding agent receives the same execution traces, host tool inventory, interface specification, and tool implementation constraints as the full run. Both FindingDory configurations use three distillation steps: capability ledger and gap map, tool synthesis with code self-tests, and skill writing. It receives no exploration programs, notes, answers, or session records, so its flags contain no answer disagreements. In this configuration, directly viewing memory images or reading memory annotation files is prohibited. Its code may process the self-test recordings, and the agent may inspect the resulting text diagnostics and tool outputs to debug the tools; transcripts are audited for compliance. The distilled tools and skill are mounted and evaluated with the protocol above.

The run produces four tools: an event index, a revisit roster, a multi-frame zoom, and a place timeline. The event index segments intervals in which the robot base stops moving. These stops follow placements, so the index records one interval per interaction and no pickup. It reports 9, 9, and 11 interactions in three of the ten episodes, whereas every episode contains ten and the full run's event tool finds ten in each. The count of ten interactions applies to held-out episodes 81--90, whereas the illustrated episode 98 from episodes 91--100 contains eleven pick-and-place interactions; these event counts are independent of the 15-round online decision budget. Durations are measured as the length of a stop instead of the span from pickup to placement. Table~\ref{tab:fd-wo-explore} gives success by query type; the largest gaps to full EDM are ordered revisitation, duration tracking, and interaction order. The configuration answers 110 queries that the unmounted executor misses and misses 57 that it answers (exact McNemar $p=5\times10^{-5}$); against full EDM the counts are 28 and 144 ($p=5\times10^{-20}$). Its tools return short tables and the executor submits after 4.1 rounds on average, which lowers tokens per query to 26.9k.

\begin{table}[ht]
\centering
\caption{Success (\%) on FindingDory episodes 81--90 by the benchmark's low-level query category; $n$ is the number of queries.}
\label{tab:fd-wo-explore}

\setlength{\tabcolsep}{4pt}
\begin{tabular}{@{}lrrrr@{}}
\toprule
\textbf{Query type} & $n$ & \textbf{w/o EDM} & \textbf{EDM w/o Explore} & \textbf{w/ EDM} \\
\midrule
Interaction Order & 160 & 31.9 & 46.9 & 82.5 \\
Conditional Interaction & 68 & 42.6 & 44.1 & 51.5 \\
Interaction & 60 & 68.3 & 68.3 & 73.3 \\
Unordered Revisitation & 59 & 6.8 & 18.6 & 30.5 \\
Room Visitation & 50 & 74.0 & 72.0 & 86.0 \\
Spatial Relationship & 50 & 4.0 & 8.0 & 12.0 \\
Object Attributes & 50 & 52.0 & 64.0 & 68.0 \\
Ordered Revisitation & 30 & 0.0 & 40.0 & 96.7 \\
Duration Tracking & 30 & 33.3 & 36.7 & 76.7 \\
Object Recall & 20 & 45.0 & 45.0 & 45.0 \\
Time-Based & 20 & 75.0 & 80.0 & 100.0 \\
\bottomrule
\end{tabular}
\end{table}

\subsection{Real Patrol Memory}
\label{app:robot-evaluation}
\label{app:full-queries}
We use an AGIBOT A3 robot to collect RGB observations and LiDAR-based mapping poses in an office building. The evaluation uses 648 keyframes from floor 8 and 501 from floor 1. GPT-5.6-luna annotates the memories, and visual and text embeddings are computed locally. Robot motion is disabled during evaluation; answers and cited historical evidence are scored. The real-robot interface offers \texttt{semantic\_search} for English content queries (three returns by default, at most ten, with coordinates and frame paths) and \texttt{inspect\_frames} for at most three original images. EDM adds text lookup, trajectory computation, photo matching, and a skill.

Execution is limited to ten tool-interaction rounds; exploration has a separate 400-round limit. Each of six tasks runs independently five times with the same prompt and image input in both configurations. Warmth Corner, farthest point, and booth counting are exploration tasks. Room 805 is a new query on the same memory; patrol length and photo localization use the new floor-1 memory. Frozen programs are reused and, where needed, rebuild indices on floor 1.

Sign queries need a cited frame showing the text, with a one-frame tolerance. Farthest point requires frame $349\pm3$ and distance $31.07\pm1.5$\,m. Booth counting needs evidence for both instances. Patrol length succeeds within 10\% of 111\,m. Photo localization accepts a coordinate or cited camera pose within 3\,m of the reference.

These are the six actual English prompts. \texttt{[BOOTH-REF]} stands for the booth reference photo. \texttt{[PHOTO-LOC]} is an independently captured phone photograph of the floor-1 environment. Both configurations receive identical images.

\paragraph{Warmth Corner (8F).} ``Where is the Warmth Corner? Find in the memory the wall sign that reads WARMTH CORNER and give a frame\_id in which that sign is clearly legible.''
\paragraph{Farthest point (8F).} ``Taking the patrol start (the position of the first frame) as the reference, where on the whole trajectory is the point farthest from the start in straight-line distance? Give the frame\_id at that point and the straight-line distance in meters.''
\paragraph{Booth count (8F).} ``Here is a photo of a phone booth: [BOOTH-REF]. How many booths like this one did the robot see on this patrol? Give one frame\_id for each.''
\paragraph{Room 805 (8F).} ``Where does it say 805? Find in the memory the door plate or sign that reads 805 and give a frame\_id in which the number is clearly legible.''
\paragraph{Patrol length (1F).} ``How far did the robot walk in total during this patrol? Give the total walked distance in meters along the robot's own path (not a straight-line distance).''
\paragraph{Photo localization (1F).} ``Here is a photo: [PHOTO-LOC]. Where was it taken? Give the frame\_id closest to that spot and its map coordinates (x, y in meters).''

\subsection{Store Annotator Replacement}
\label{app:annotator}
For OC-NaVQA and FindingDory we replace the GPT-5.6-sol writing-time annotator with locally served Qwen3.8-27B-FP8 and rebuild text embeddings. Keyframes, prompt, parser, retrieval channels, execution model, and online limits remain fixed. Programs distilled on the Sol-annotated store are reused directly on the Qwen-annotated store. Local Qwen annotation incurs no annotation API charge.

\begin{table}[ht]
\centering
\caption{OC-NaVQA annotator replacement on 210 questions. All rows use the GPT-5.6-sol executor. Tokens/q reports mean online input-plus-output tokens in thousands. Green denotes configurations with EDM; bold marks column bests.}
\label{tab:app_annotator_oc}

\setlength{\tabcolsep}{1.5pt}
\renewcommand{\arraystretch}{1.05}
\begin{tabular}{@{}lrrrrrr@{}}
\toprule
& \multicolumn{2}{c}{\textbf{Accuracy (\%)}} & \multicolumn{2}{c}{\textbf{Position}} & \textbf{Temporal} & \\
\cmidrule(lr){2-3}\cmidrule(lr){4-5}
\textbf{Annotator / configuration} & \textbf{Binary} $\uparrow$ & \textbf{Text} $\uparrow$ & \textbf{Mean (m)} $\downarrow$ & $\le$\textbf{10\,m (\%)} $\uparrow$ & \textbf{Error (min)} $\downarrow$ & \shortstack[r]{\textbf{Tokens/q}\\\textbf{(k)} $\downarrow$} \\
\midrule
GPT-5.6-sol (w/o EDM) & \textbf{86.6} & 63.6 & 33.10 & 19.7 & 0.587 & \textbf{18.8} \\
\rowcolor{resightTint}
GPT-5.6-sol (w/ EDM) & 82.1 & 63.6 & \textbf{24.72} & \textbf{36.6} & 0.475 & 20.4 \\
\addlinespace[2pt]
Qwen3.8-27B (w/o EDM) & 85.1 & \textbf{69.7} & 31.54 & 16.9 & \textbf{0.427} & 19.1 \\
\rowcolor{resightTint}
Qwen3.8-27B (w/ EDM) & 85.1 & \textbf{69.7} & 28.29 & 35.2 & 0.463 & 21.3 \\
\bottomrule
\end{tabular}
\end{table}

\begin{table}[ht]
\centering
\caption{FindingDory annotator replacement on episodes 91--100 (596 queries). High-level success (\%); all rows use the GPT-5.6-sol executor. Tokens/q reports mean online input-plus-output tokens in thousands. Green denotes configurations with EDM; bold marks column bests.}
\label{tab:app_annotator}

\setlength{\tabcolsep}{3pt}
\renewcommand{\arraystretch}{1.05}
\begin{tabular}{@{}lrrrrr@{}}
\toprule
\textbf{Annotator / configuration} & \textbf{Spatial} $\uparrow$ & \textbf{Temporal} $\uparrow$ & \textbf{Multi-goal} $\uparrow$ & \textbf{Overall} $\uparrow$ & \shortstack[r]{\textbf{Tokens/q}\\\textbf{(k)} $\downarrow$} \\
\midrule
GPT-5.6-sol (w/o EDM) & 45.5 & 41.0 & 7.9 & 38.3 & 130 \\
\rowcolor{resightTint}
GPT-5.6-sol (w/ EDM) & \textbf{54.2} & 82.9 & \textbf{50.6} & 63.8 & 43 \\
\addlinespace[2pt]
Qwen3.8-27B (w/o EDM) & 48.8 & 37.6 & 7.9 & 38.8 & 144 \\
\rowcolor{resightTint}
Qwen3.8-27B (w/ EDM) & \textbf{54.2} & \textbf{87.1} & \textbf{50.6} & \textbf{65.3} & \textbf{42} \\
\bottomrule
\end{tabular}
\end{table}

On either annotated store, EDM improves all three FindingDory categories.

\section{Compute Cost and Online Usage}
\label{app:budgets}

\subsection{Offline Construction and Capability Expansion}
Keyframe annotation, vector encoding, and EDM capability construction are performed offline and can serve multiple queries. Appendix~\ref{app:annotator} compares full task results for Sol and locally annotated Qwen stores. Figures~\ref{fig:app-tokens} and~\ref{fig:app-requests} report online execution costs.

\subsection{Online Tokens per Query}
\label{app:online-tokens}
\begin{figure}[t]
\centering
\includegraphics[width=\linewidth]{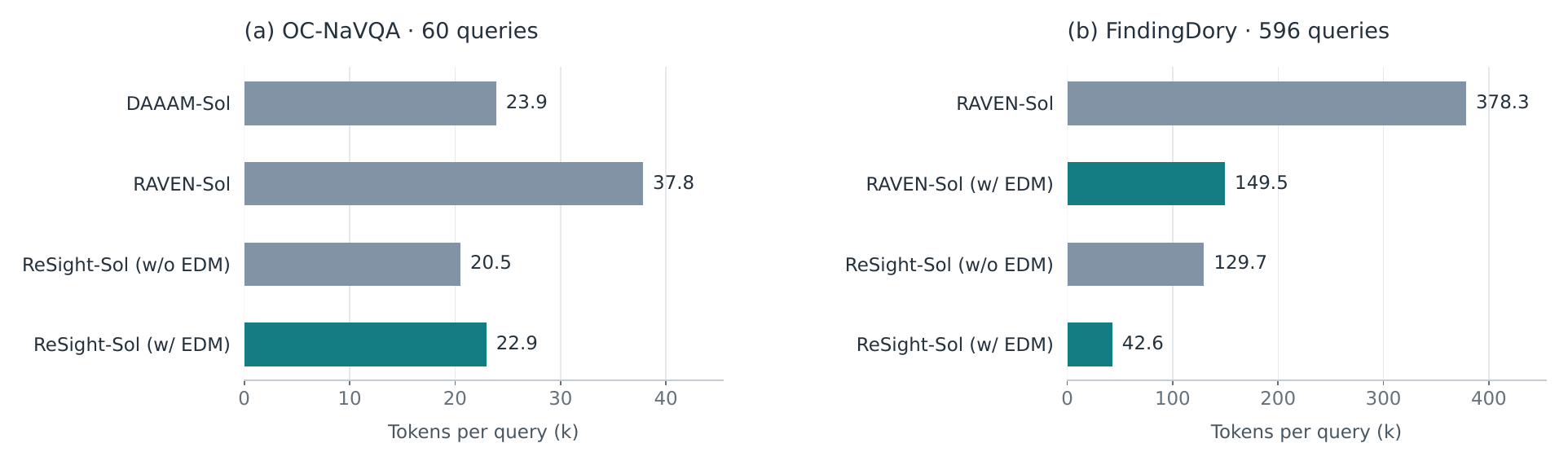}
\caption{Mean online tokens per query (thousands). Gray denotes the base host and green its EDM-mounted counterpart. The two benchmarks use separate horizontal scales.}
\label{fig:app-tokens}
\end{figure}

Figure~\ref{fig:app-tokens} sums input and output tokens across online model requests, including context carried into later requests. OC-NaVQA usage is measured on the 60-query subset from sequences 0 and 21, with all four systems calling GPT-5.6-sol through Azure. FindingDory uses all 596 questions and one deployment across its four configurations.

On FindingDory, EDM reduces online tokens on both hosts; the ReSight executor uses slightly more tokens on the OC subset. GOAT-core retrieval-only makes no online model request during candidate selection, while the agent uses 6.7k tokens per query on average.

\subsection{Online Model Requests per Query}
\label{app:model-requests}
\begin{figure}[t]
\centering
\includegraphics[width=\linewidth]{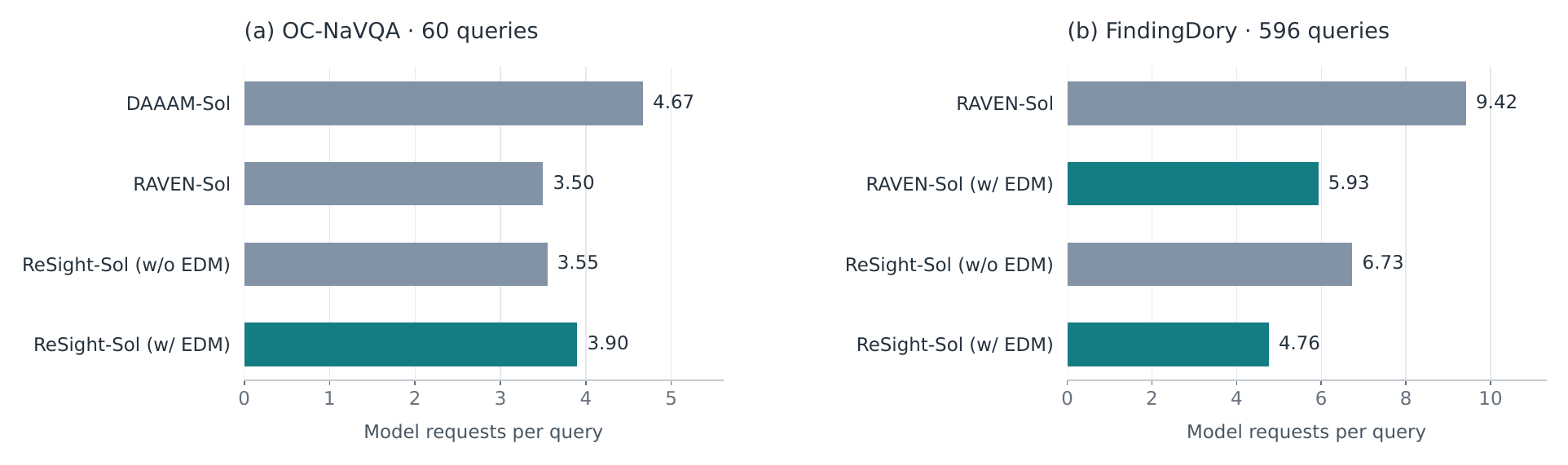}
\caption{Mean online model requests per query. ReSight includes decision and finalization requests; RAVEN includes agent decisions and answer generation; DAAAM uses step-level request logs.}
\label{fig:app-requests}
\end{figure}

Figure~\ref{fig:app-requests} counts every online executor model request: decisions to retrieve or inspect, answer submission, and any additional finalization. Multiple tool calls emitted by one model response count once. The task sets match Figure~\ref{fig:app-tokens}.

Both FindingDory hosts make fewer requests after mounting. ReSight makes slightly more on the OC subset. The decision limits for each host appear in Appendix~\ref{app:ocnavqa}.

\section{Complete Execution Trajectories}
\label{app:trajectories}
\label{app:case-records}
We present paired execution trajectories before and after EDM for two OC-NaVQA queries, two FindingDory queries, and six real-robot task types. Each case shows the model requests, tool calls, selected returns, visual evidence, and final answers. We then demonstrate how memory-query results support subsequent physical execution.

\begingroup
\raggedbottom
\clearpage
\subsection{OC-NaVQA}
\label{app:trajectory-oc}
\begin{figure}[H]
\centering
\includegraphics[width=0.98\textwidth]{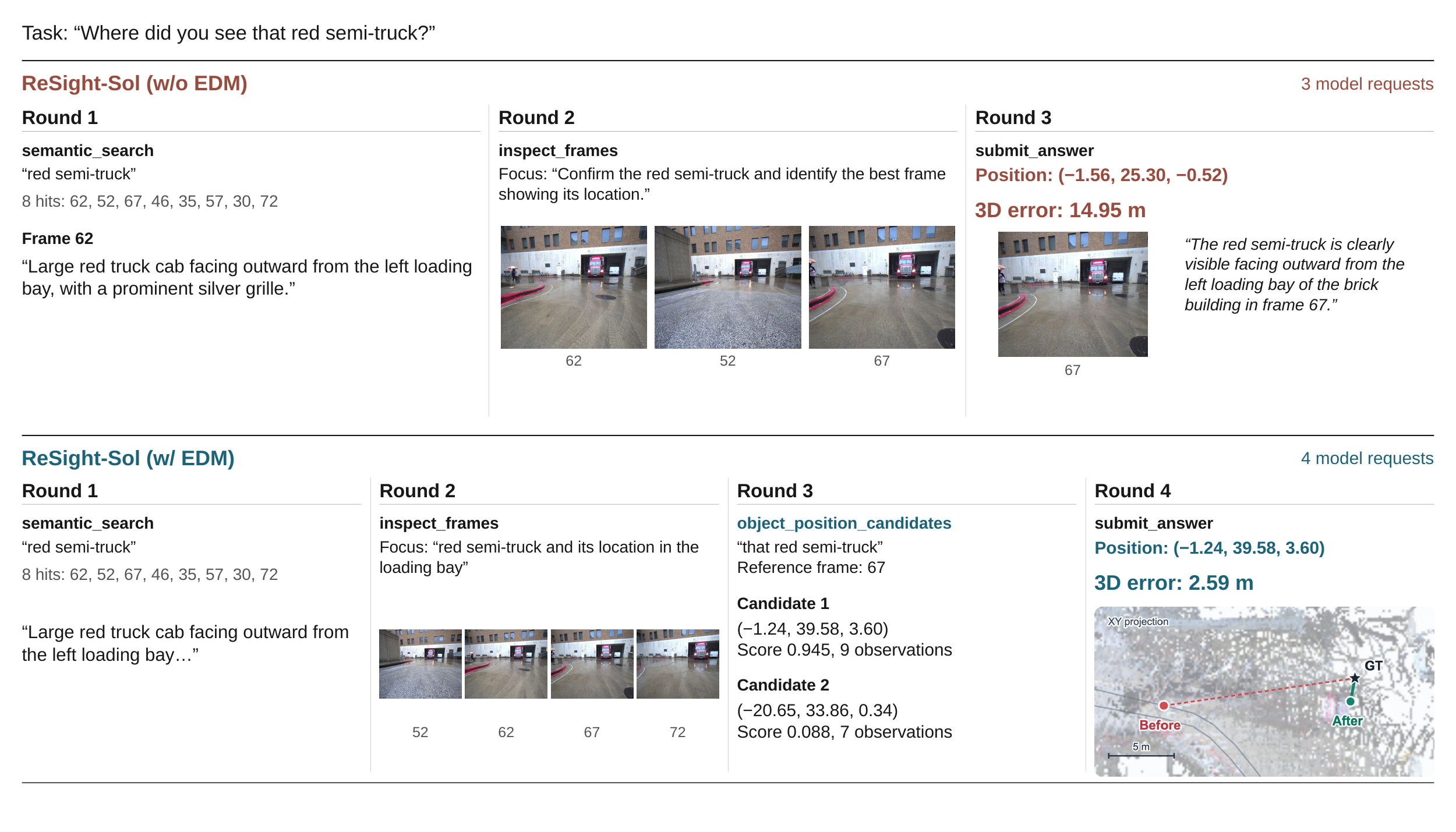}
\caption{OC-NaVQA: red semi-truck. Both executions cite frame 67; EDM uses several views to estimate the truck's position rather than returning the camera pose.}
\label{fig:traj-oc-truck}
\end{figure}

\begin{figure}[H]
\centering
\includegraphics[width=0.98\textwidth]{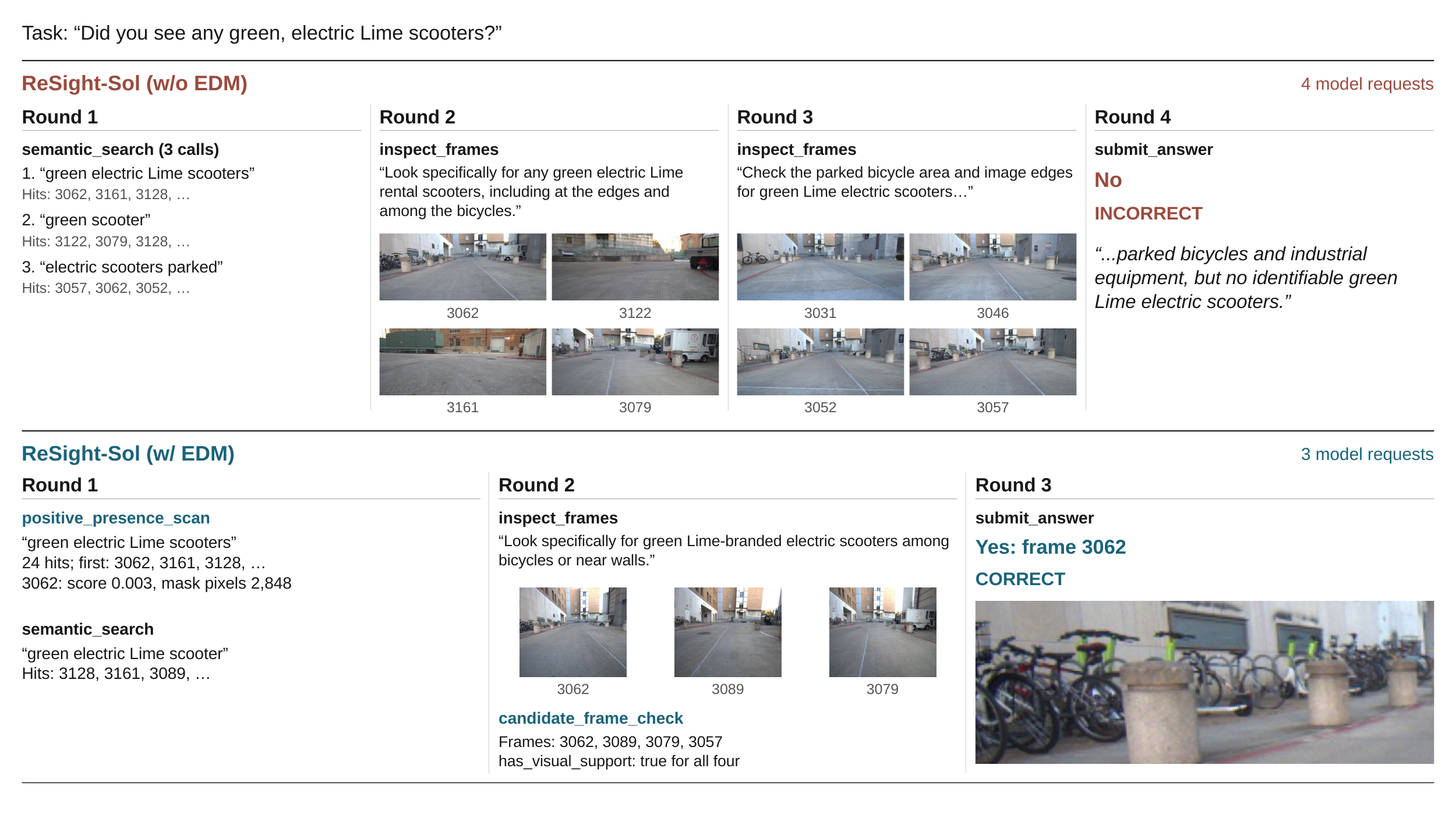}
\caption{OC-NaVQA: green Lime scooters. The base execution misses the target. EDM's presence scan narrows the candidate frames for image verification.}
\label{fig:traj-oc-scooters}
\end{figure}

\clearpage
\subsection{FindingDory}
\label{app:trajectory-fd}
\begin{figure}[H]
\centering
\includegraphics[width=0.98\textwidth]{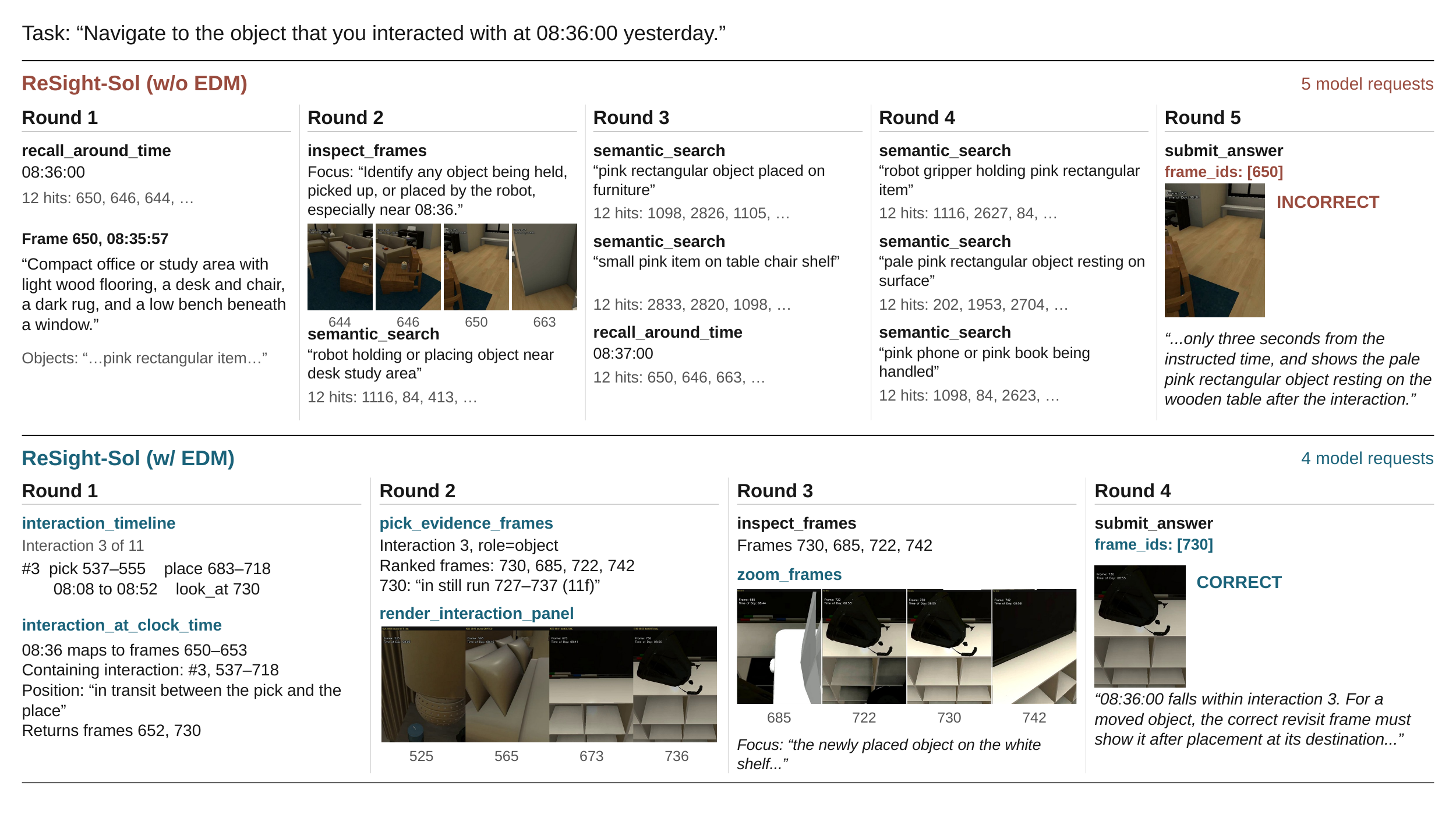}
\caption{FindingDory: interaction time. The base execution cites a frame before placement; EDM identifies the interaction and cites evidence after placement.}
\label{fig:traj-fd-time}
\end{figure}

\begin{figure}[H]
\centering
\includegraphics[width=0.98\textwidth]{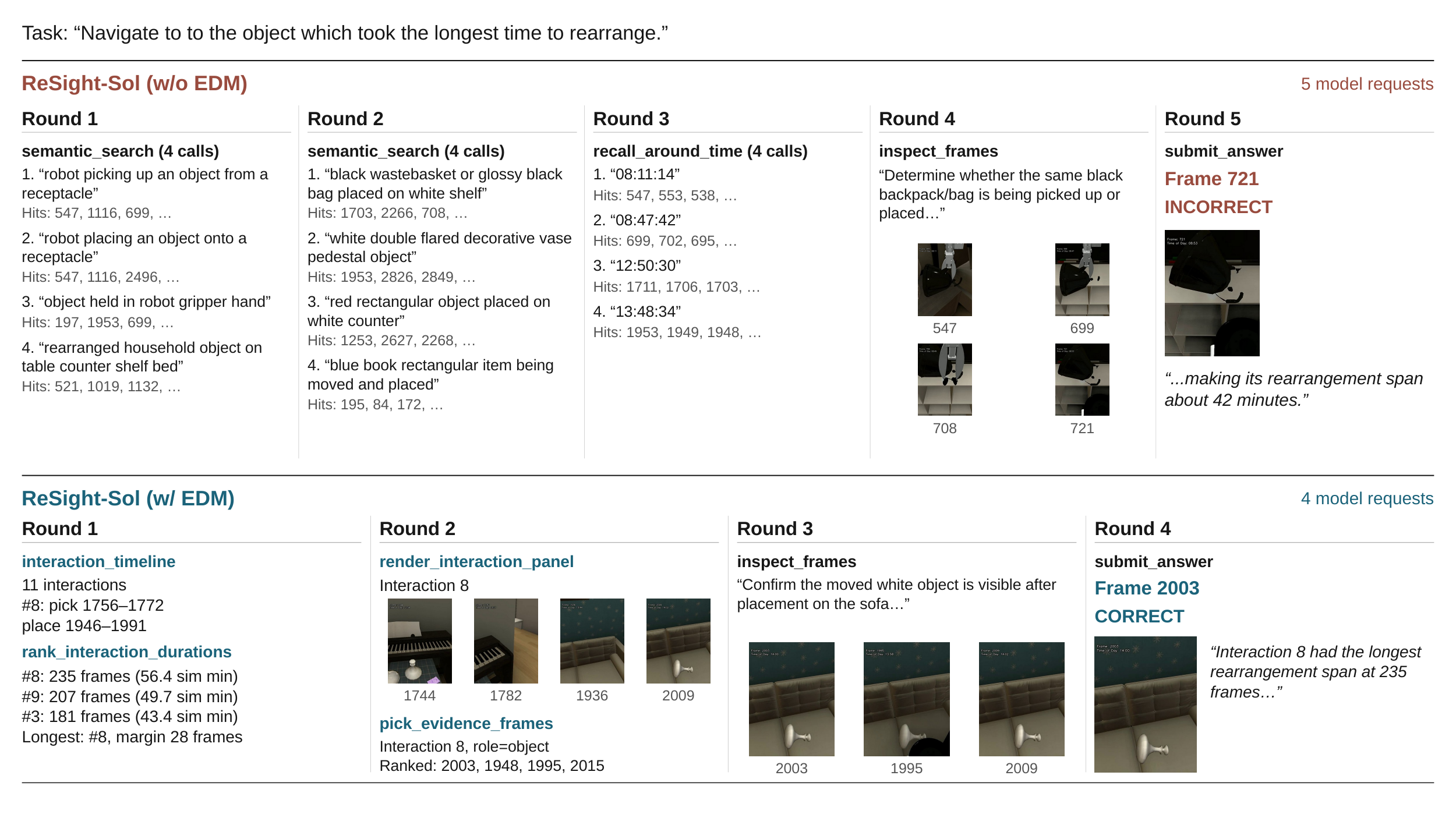}
\caption{FindingDory: the object with the longest rearrangement. The base execution estimates duration from retrieved frames; EDM ranks indexed interactions by duration and checks the selected object's evidence. The repeated word ``to'' is retained from the original benchmark question.}
\label{fig:traj-fd-longest}
\end{figure}

\clearpage
\subsection{Real Robot Patrol Memory}
\label{app:trajectory-robot}
\begin{figure}[H]
\centering
\includegraphics[width=0.98\textwidth]{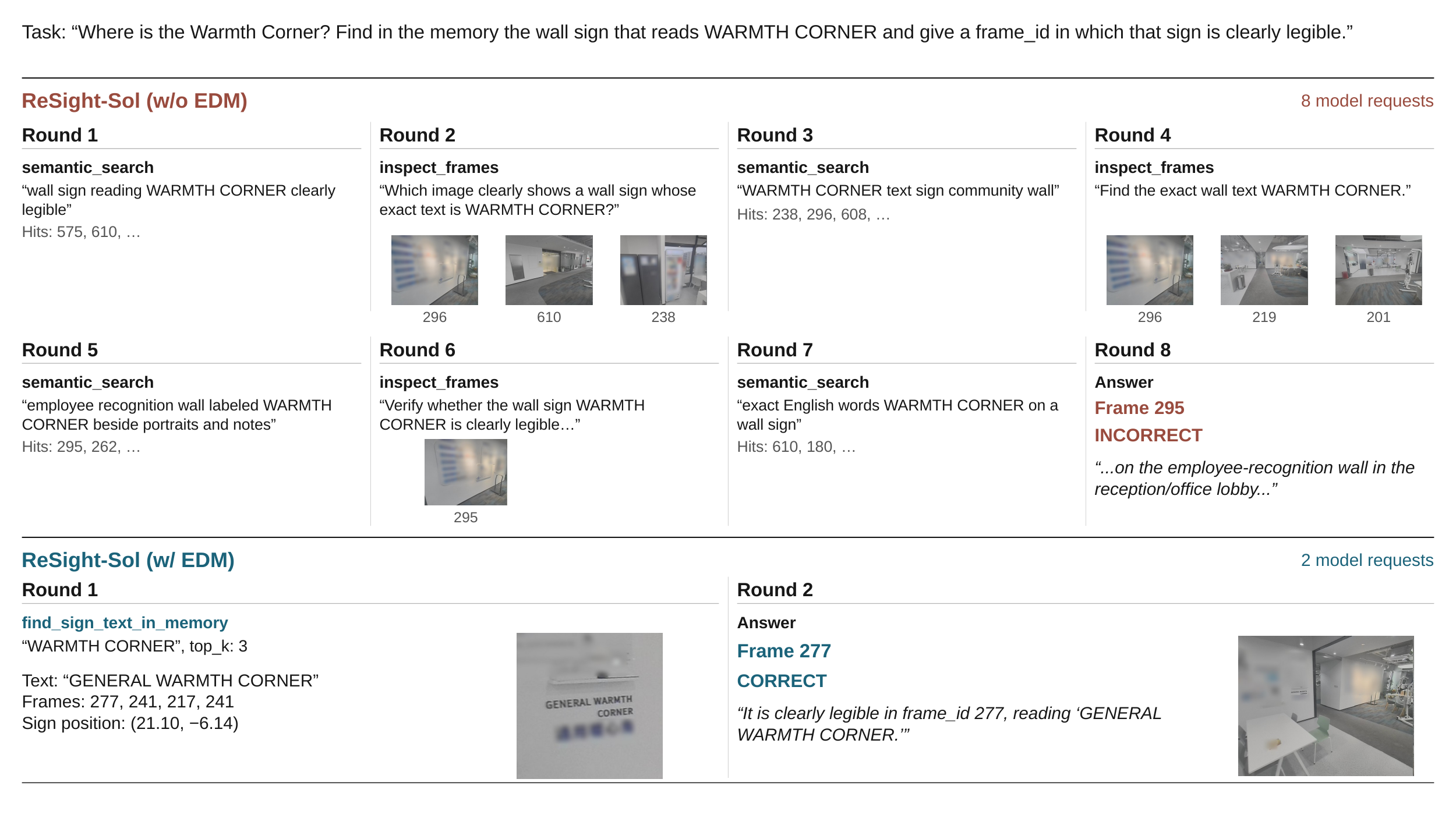}
\caption{Robot patrol, held-in: Warmth Corner sign. The base execution selects an unrelated recognition wall; text lookup returns a frame with the legible sign.}
\label{fig:traj-robot-warmth}
\end{figure}

\begin{figure}[H]
\centering
\includegraphics[width=0.98\textwidth]{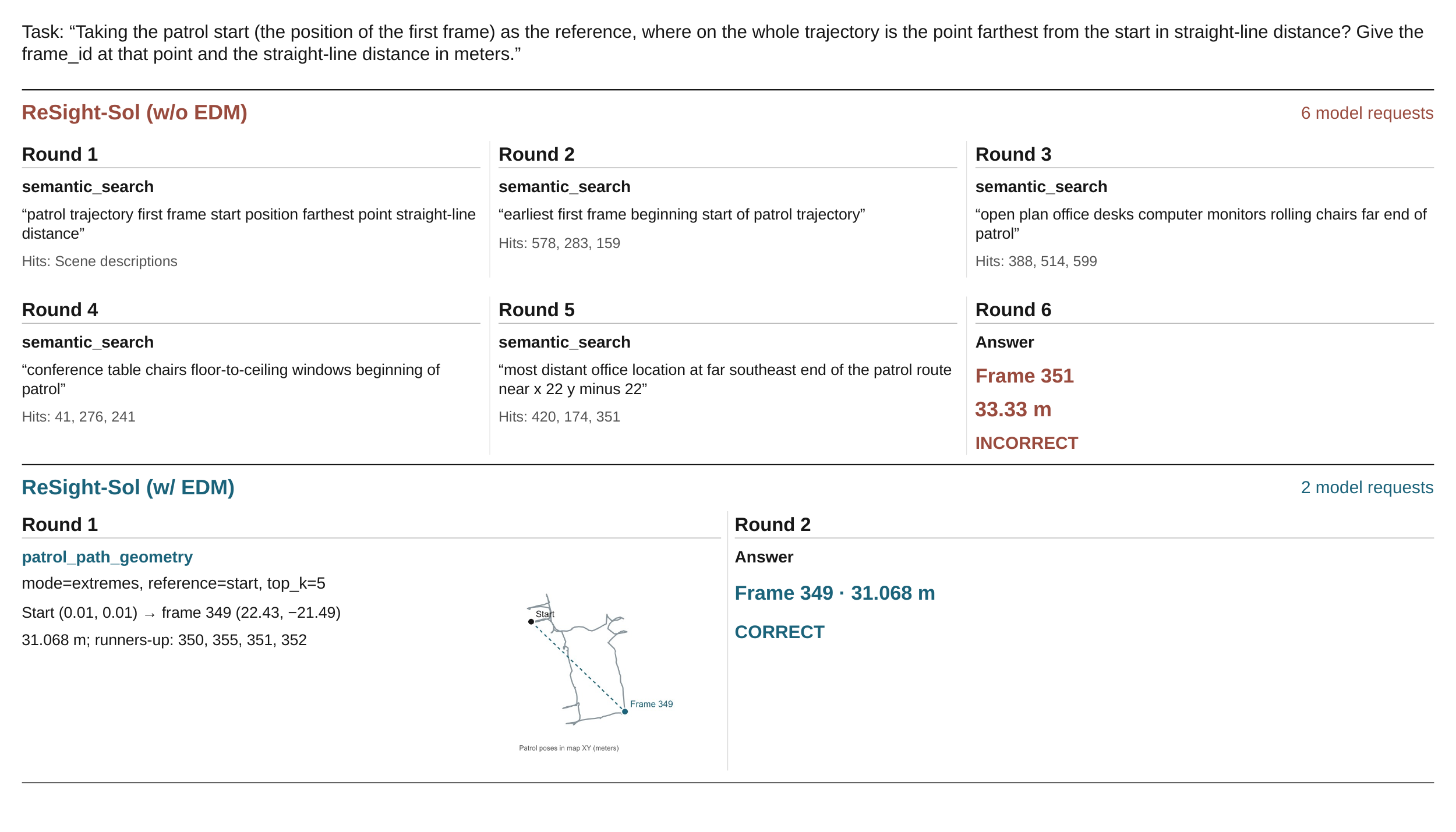}
\caption{Robot patrol, held-in: farthest point. EDM computes the farthest frame from all 648 stored poses. The plotted trajectory is reconstructed from stored poses.}
\label{fig:traj-robot-farthest}
\end{figure}

\begin{figure}[H]
\centering
\includegraphics[width=0.98\textwidth]{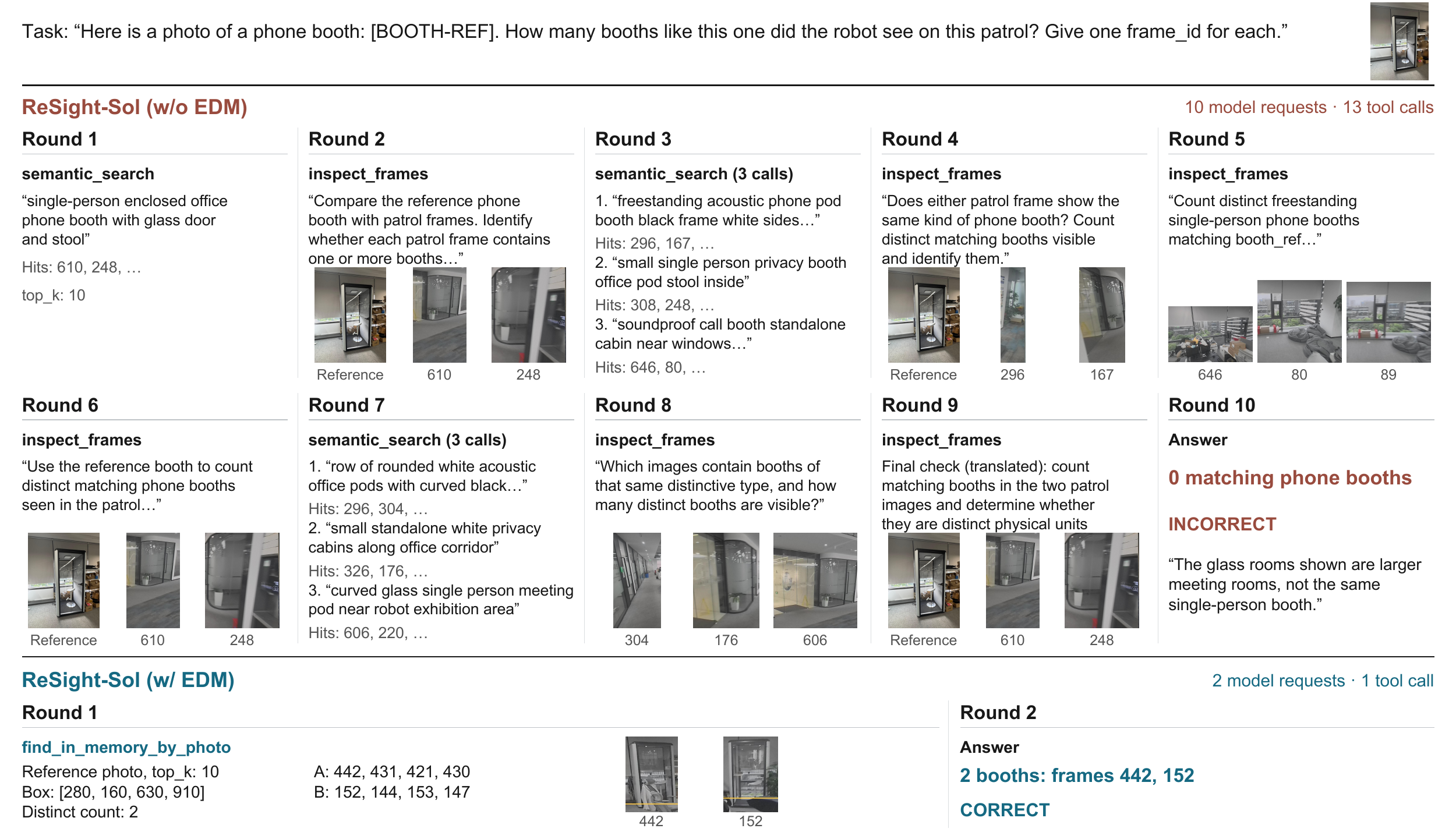}
\caption{Robot patrol, held-in: booth count. Under the same short prompt and reference photo, the base execution answers zero, while EDM identifies both booths (frames 442 and 152). }
\label{fig:traj-robot-booths}
\end{figure}

\begin{figure}[H]
\centering
\includegraphics[width=0.98\textwidth]{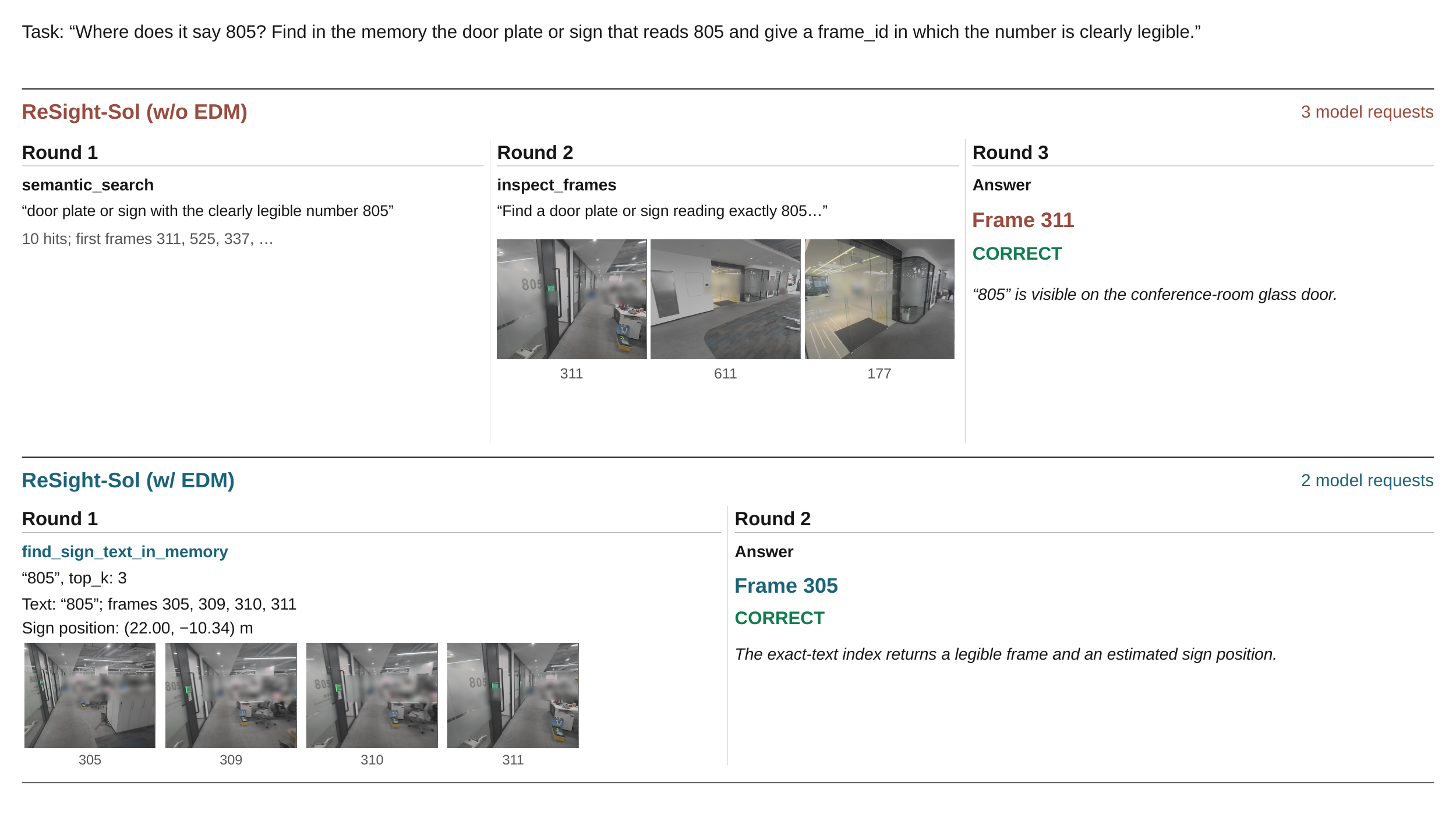}
\caption{Robot patrol, held-out: room 805 sign. Both executions answer correctly; the mounted text index reduces the count from three model requests to two.}
\label{fig:traj-robot-805}
\end{figure}

\begin{figure}[H]
\centering
\includegraphics[width=0.98\textwidth]{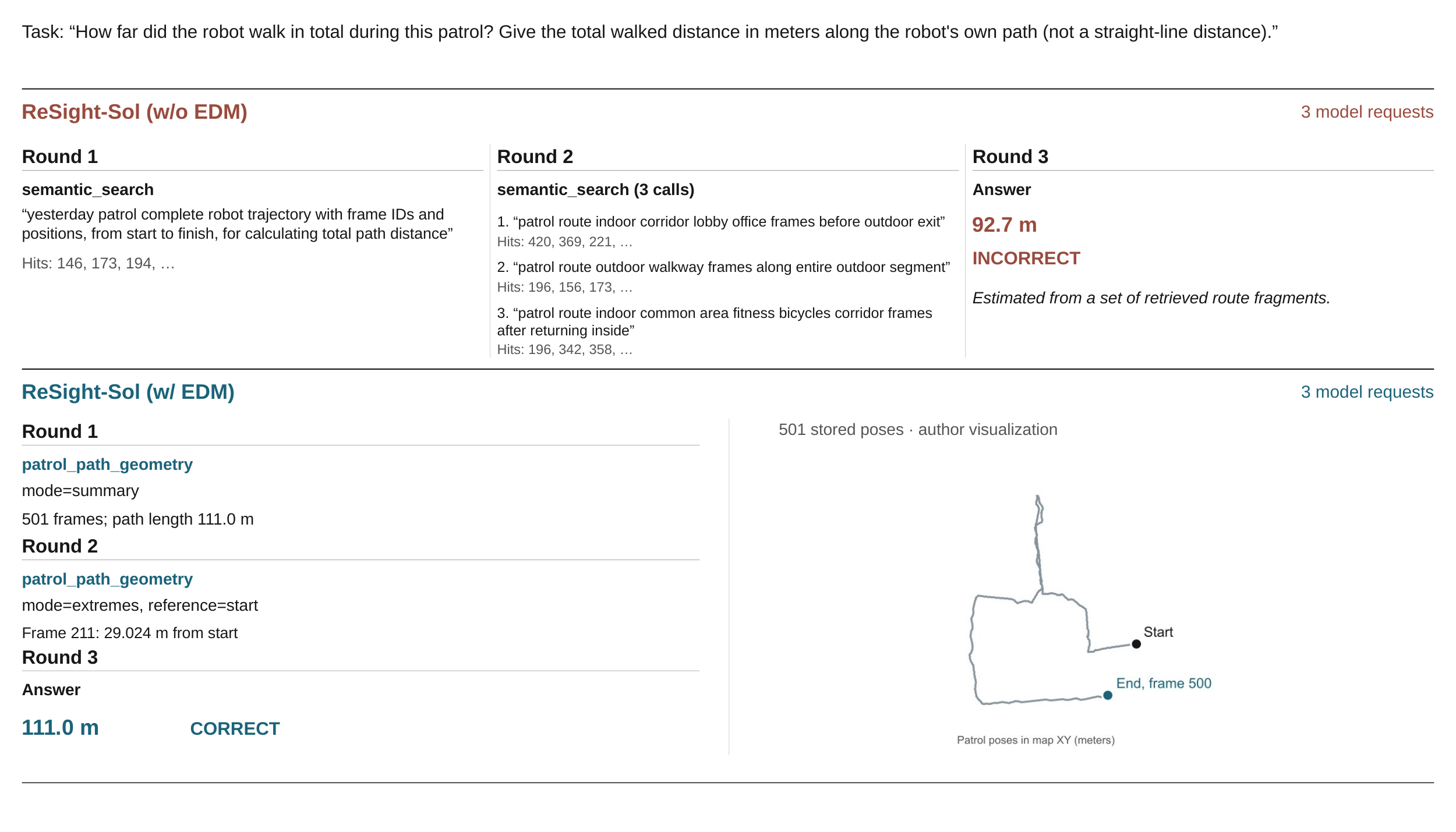}
\caption{Robot patrol, held-out: total distance walked. The base execution estimates 92.7\,m from route fragments; EDM sums consecutive positions across 501 frames to obtain 111.0\,m. The plotted trajectory is reconstructed from stored poses.}
\label{fig:traj-robot-distance}
\end{figure}

\begin{figure}[H]
\centering
\includegraphics[width=0.98\textwidth]{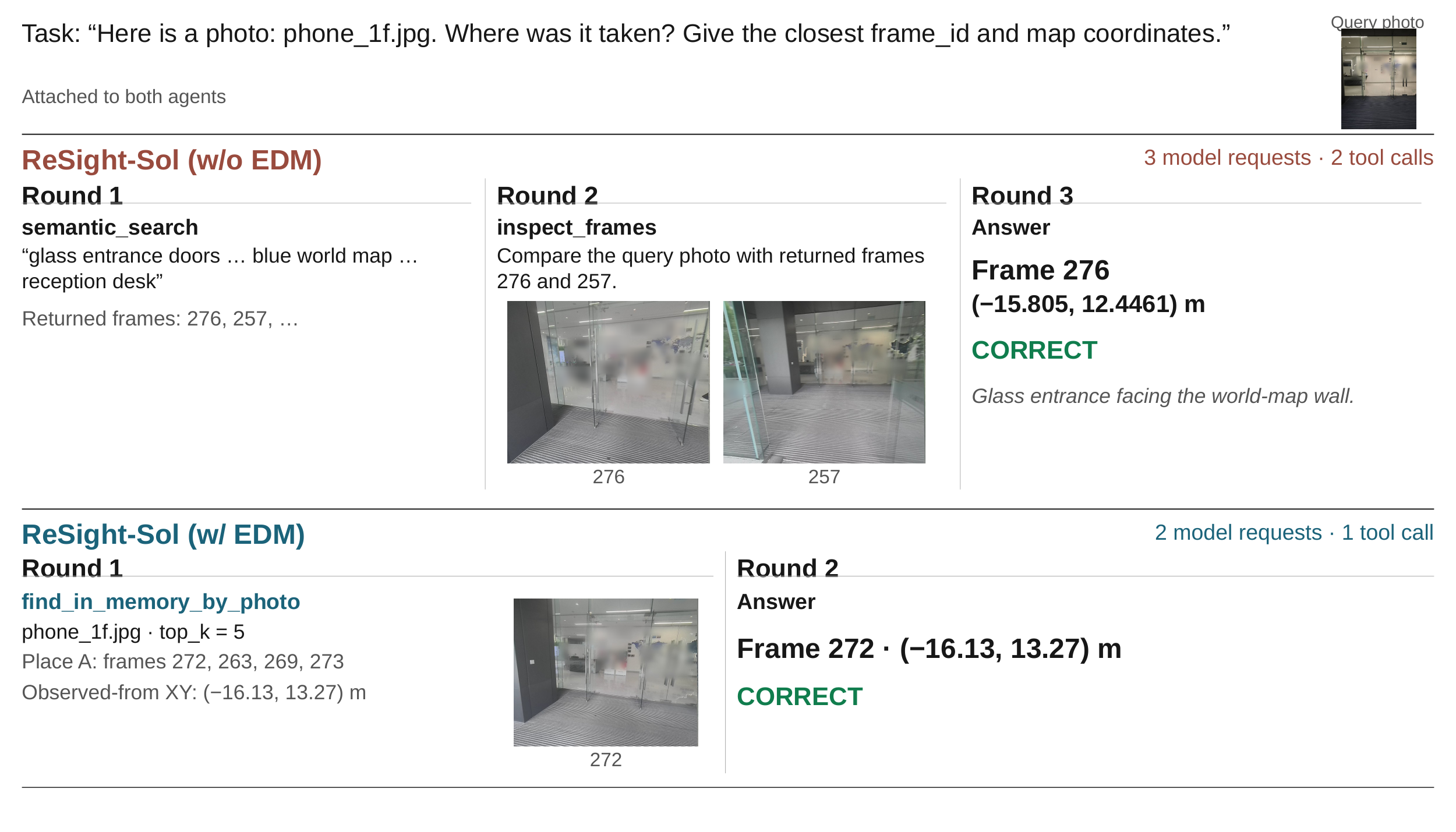}
\caption{Robot patrol, held-out: independent phone-photo localization. Both executions answer correctly. In this paired run, EDM uses one tool call and two model requests, versus two tool calls and three model requests without EDM.}
\label{fig:traj-robot-photo}
\end{figure}
\clearpage

\subsection{From Memory Queries to Physical Execution}
\label{app:robot-execution}
The following demonstrations continue the booth-counting and photo-localization queries with physical actions requested by the user. Retrieved locations guide navigation, while fresh observations support on-site verification.

\begin{figure}[H]
\centering
\includegraphics[width=\textwidth]{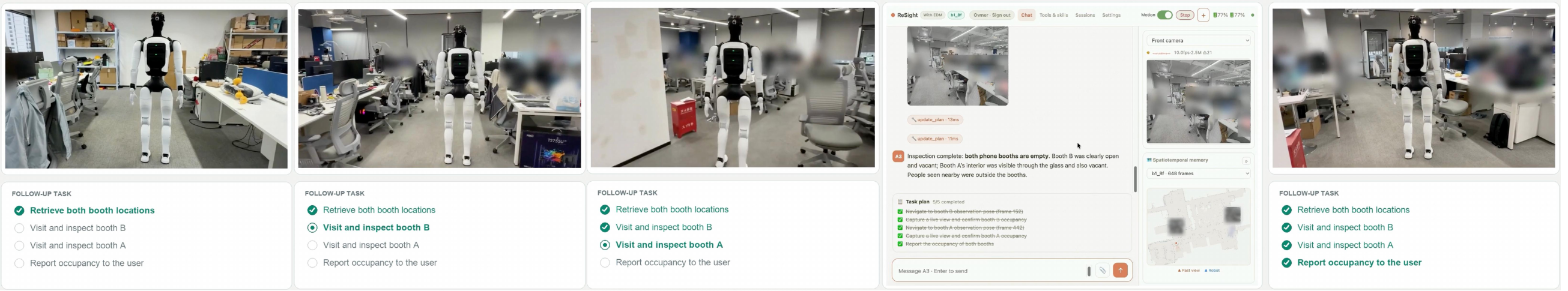}
\caption{Visiting and inspecting two phone booths on floor 8. After the memory query identifies two booths, the user asks: ``Help me plan a route to visit these two phone booths right now and confirm in person whether anyone is there.'' The robot visits booth B and then booth A, checks their current occupancy, and reports that both are empty.}
\label{fig:robot-booth-navigation}
\end{figure}

\begin{figure}[H]
\centering
\includegraphics[width=\textwidth]{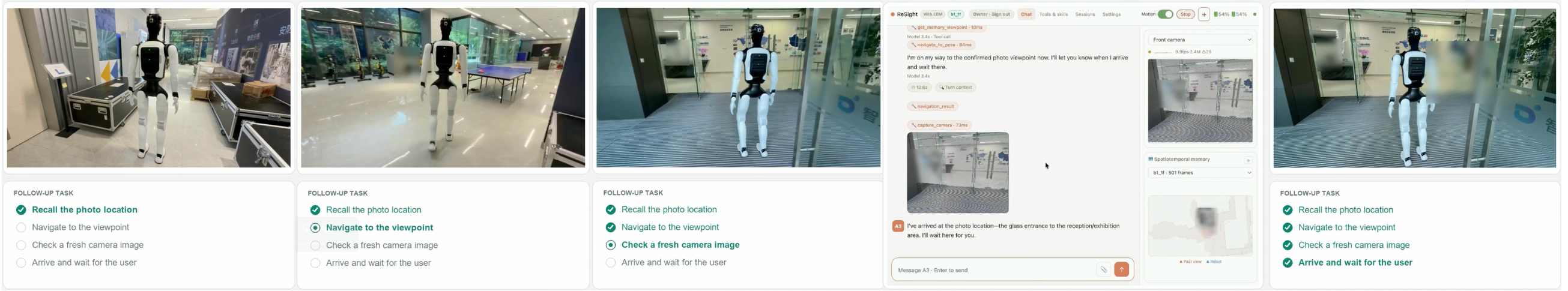}
\caption{Navigating to a photo-matched location on floor 1. After the memory query locates the reference photograph, the user asks: ``Go there and wait for me.'' The robot navigates to the retrieved viewpoint, checks a fresh camera image, and waits at the destination.}
\label{fig:robot-photo-navigation}
\end{figure}
\clearpage

\endgroup

\end{document}